\documentclass[lettersize,journal]{IEEEtran}
\usepackage[utf8]{inputenc}
\usepackage{amsmath,amsfonts,amssymb}
\usepackage{algorithmic}
\usepackage{algorithm}
\usepackage[T1]{fontenc}    
\usepackage{bm} 

\makeatletter
\newcommand{\HYPERPARAMS}{\item[\textbf{Hyperparameters:}]}
\makeatother
\usepackage{array}
\usepackage{multirow}
\usepackage{tabularx}
\usepackage{arydshln}
\usepackage{booktabs}
\usepackage{graphicx}
\usepackage{float}
\usepackage{epstopdf}
\usepackage[caption=false,font=normalsize,labelfont=sf,textfont=sf]{subfig}
\usepackage{textcomp}
\usepackage{verbatim}
\usepackage{pifont}
\usepackage{stfloats}
\usepackage{color}
\usepackage{dsfont}
\usepackage{url}
\usepackage{cite}

\begin{document}

\title{Interpretable Maximum Margin Deep Anomaly Detection}

\author{Zhiji~Yang , \IEEEmembership{Member, IEEE} ,
        Mei~Huang,
        Xinyu~Li,
        Xianli~Pan,
        Qi~Wang, Jianhua Zhao, \IEEEmembership{Senior Member, IEEE}

\thanks{Manuscript received XXX, 2026; revised XXX, 2026. This work was supported in part by the National Natural Science Foundation of China under Grant 62006206, 12161089; in part by Beijing Natural Science Foundation under Grant Z240004; in part by the Scientific Research Fund Project of Yunnan Provincial Department of Science and Technology under Grant 202501AT070460. (Corresponding authors: Xianli~Pan; Jianhua Zhao.)}

\thanks{Zhiji~Yang, Mei~Huang, Xinyu~Li, Qi~Wang, and Jianhua Zhao are with the School of Statistics and Mathematics, Yunnan University of Finance and Economics, Kunming 650221, China (e-mail: yangzhiji@ynufe.edu.cn; jhzhao.ynu@gmail.com).}
\thanks{Xianli~Pan is with the Academy for Multidisciplinary Studies, Capital Normal University, Beijing 100048, China (e-mail: pxlcjs@126.com).}
}

\markboth{}%
{Yang \MakeLowercase{\textit{et al.}}: Interpretable Maximum Margin Deep Anomaly Detection}

\maketitle

\begin{abstract}
Anomaly detection is a crucial machine-learning task with wide-ranging applications. Deep Support Vector Data Description (Deep SVDD) is a prominent deep one-class method, but it is vulnerable to hypersphere collapse, often relies on heuristic choices for hypersphere parameters, and provides limited interpretability. To address these issues, we propose Interpretable Maximum Margin Deep Anomaly Detection (IMD-AD), which leverages a small set of labeled anomalies and a maximum margin objective to stabilize training and improve discrimination. It is inherently resilient to hypersphere collapse. Furthermore, we prove an equivalence between hypersphere parameters and the network's final-layer weights, which allows the center and radius to be learned end-to-end as part of the model and yields intrinsic interpretability and visualizable outputs. We further develop an efficient training algorithm that jointly optimizes representation, margin, and final-layer parameters. Extensive experiments and ablation studies on image and tabular benchmarks demonstrate that IMD-AD empirically improves detection performance over several state-of-the-art baselines while providing interpretable decision diagnostics.
\end{abstract}


\section{Introduction}
\label{ref1}
\IEEEPARstart{A}{nomaly} detection refers to the identification of samples in a dataset that deviate significantly from expected patterns \cite{chandola_anomaly_2009}. It is widely applied in diverse domains such as liquid rocket engine \cite{9755996}, fault diagnosis \cite{9780551}, and pavement crack detection \cite{10765917}.

Anomaly detection methods can generally be divided into two categories: classical approaches and deep learning-based approaches. Classical approaches include distance-based\cite{10492459}, statistics-based \cite{9927321}, density-based \cite{9178989} and classification-based methods \cite{scholkopf1999support,tax2004support, wu2009small}. These methods are simple and provide interpretability and visualization, which facilitate the identification of abnormal patterns. However, their reliance on manual feature engineering makes them inadequate for high-dimensional or complex data in the era of big data. This limitation has motivated the development of deep anomaly detection techniques that automatically learn discriminative representations via neural networks \cite{10612766}. Existing deep approaches can be broadly grouped into three categories: reconstruction-based \cite{an2015variational,guo2024recontrast,9059022}, probabilistic-based \cite{zong2018deep}, and classification-based methods \cite{ruff2018deep,9632278,10906639}. Reconstruction-based methods assume that normal data can be accurately reconstructed while anomalies cannot. Probabilistic-based methods rely on distributional assumptions of normal data. Classification-based methods focus on learning decision boundaries around normal samples, which is advantageous when dealing with complex data distributions or limited sample sizes. % density-based \cite{kriegel2009loop},\cite{irsoy2019continuously}

Among classification-based approaches, Deep Support Vector Data Description (Deep SVDD) is a widely used formulation that replaces kernel feature maps with deep neural representations \cite{ruff2018deep, tax2004support}. Despite its advantages, Deep SVDD faces three main challenges. 

First, Deep SVDD often suffers from hypersphere collapse: training only on normal samples combined with highly expressive feature mappings may map all inputs to a single point, yielding a degenerate classifier \cite{goyal2020drocc, hojjati2024dasvdd}. Prior remedies follow two lines: objective refinements that add reconstruction or regularization terms to prevent collapse (e.g., DASVDD \cite{hojjati2024dasvdd}, DOHSC \cite{zhang2024deep}), and negative sample strategies that synthesize or exploit near-manifold negatives to tighten the boundary (e.g., DROCC \cite{goyal2020drocc}, PLAD \cite{NEURIPS2022_5c261ccd}). These methods mitigate collapse but typically introduce extra objectives or adversarial synthesis steps and do not by themselves guarantee improved parameter interpretability.

\begin{figure}[!t]
\centering
\subfloat[Deep SVDD]{\includegraphics[width=0.5\textwidth]{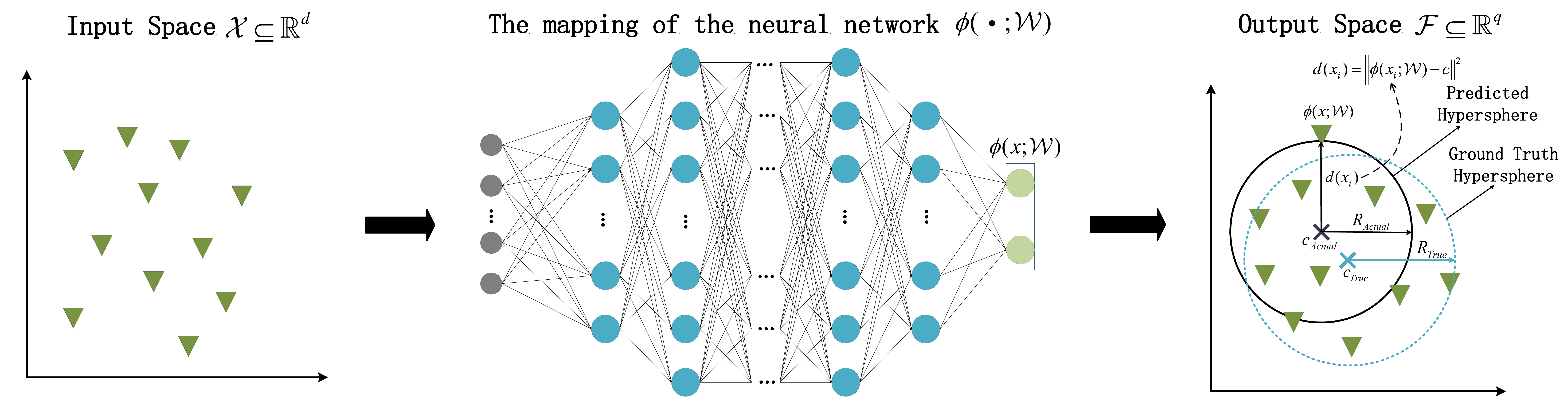}%
\label{DeepSVDDfig}}
\hfil
\subfloat[IMD-AD]{\includegraphics[width=0.5\textwidth]{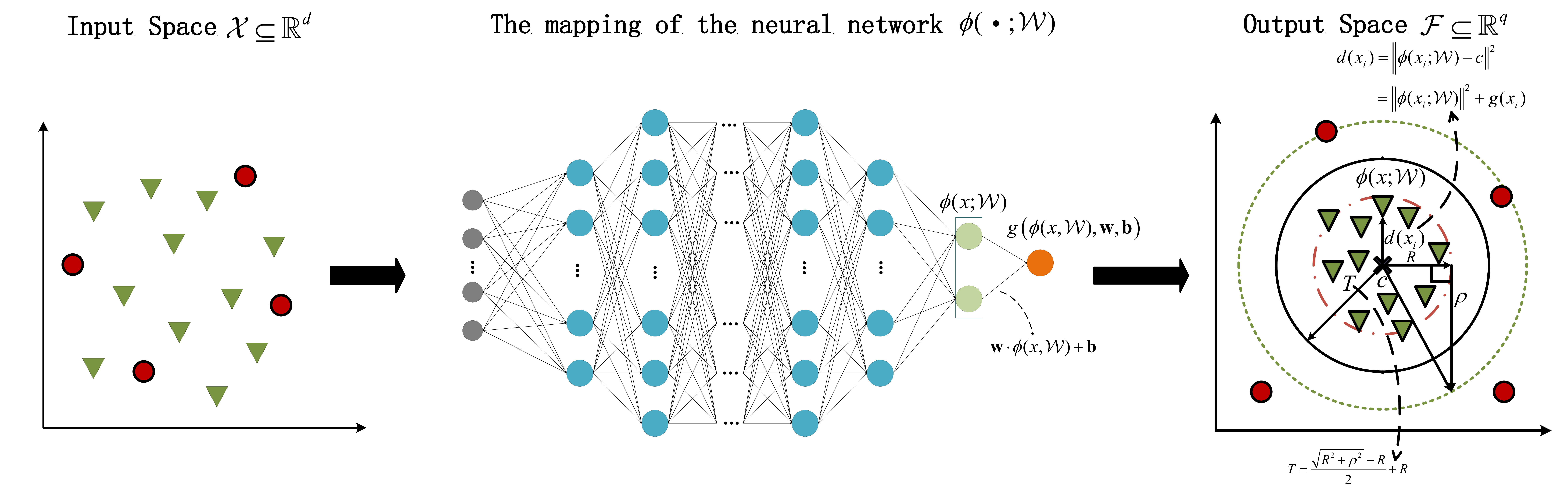}%
\label{ADMASfig}}
\caption{Training framework of Deep SVDD and IMD-AD, respectively. Green points represent normal samples, and red points represent abnormal samples. $R$ and $\mathbf{c}$ represent the radius and center of the hypersphere, respectively. $\rho$ represents the margin between abnormal samples and the surface of the hypersphere.}
\label{fig2}
\end{figure}

Second, parameter estimation (center and radius) can be inaccurate: in practice the hypersphere center and radius are often set by heuristics (e.g., initialization averages or quantiles) rather than learned jointly, which harms detection accuracy \cite{9059022,9422197}. As illustrated in the Fig. \ref{DeepSVDDfig}, such a scenario may occur in Deep SVDD: the ground trueth center and radius of the hypersphere are shown in blue, while the predicted results are depicted in black. There is a significant difference between the predicted results and the ground truth. Some works learn centers or layerwise centers to improve robustness (e.g., DASVDD \cite{hojjati2024dasvdd}, MOCCA \cite{9640579}), and DOHSC seeks better geometric alignment via orthogonality constraints \cite{zhang2024deep}. Nevertheless, a unified, provably interpretable parameterization that enables end-to-end learning of hypersphere parameters remains lacking.

Third, interpretability is limited: although post‑hoc and intrinsically interpretable methods (e.g., FCDD \cite{liznerski2021explainable}, DTD \cite{kauffmann_towards_2020}, Dual-SVDAE \cite{zhang2021deep}) provide visual explanations or heatmaps, they typically do not integrate collapse prevention and principled hypersphere parameter learning in a single end-to-end framework.

In summary, existing approaches address subsets of these problems but do not provide a cohesive solution that (i) prevents hypersphere collapse, (ii) learns hypersphere parameters in a unified, end-to-end manner with clear interpretability, and (iii) yields actionable visual explanations. To bridge this gap, we propose Interpretable Maximum Margin Deep Anomaly Detection (IMD AD), which simultaneously targets these three challenges. The contributions of this work are as follows.

\begin{enumerate}
	\itemsep=0pt
\item \textbf{Methodology for anomaly detection:} We demonstrate that our deep anomaly detection can be effectively formulated within a maximum-margin framework. By leveraging a small number of abnormal samples and explicitly maximizing the margin between normal and abnormal data, IMD-AD achieves inherent robustness against hypersphere collapse.

\item \textbf{Efficient end-to-end optimization:} We introduce an end-to-end optimization algorithm where the hypersphere center and radius are embedded as network parameters and updated directly through backpropagation. This approach removes the need for heuristic parameter tuning, thereby enhancing both computational efficiency and solution quality.

\item \textbf{Theoretical guarantee and interpretability:} We establish a theoretical equivalence between the hypersphere parameters and the final-layer weights of the neural network, providing clear interpretability and visualization of model outputs. Additionally, we analyze the interpretable properties of hyperparameters, which supports principled hyperparameter selection.
\end{enumerate}
The remainder of this paper is organized as follows. Section \ref{ref2} reviews related work. Section \ref{ref3} introduces the proposed IMD-AD method. Section \ref{ref4} provides theoretical analyses. Section \ref{ref5} reports experimental evaluations. Section \ref{ref6} concludes the paper with future research directions.

\section{Review of Deep Support Vector Data Description}
\label{ref2}
Support Vector Data Description (SVDD) \cite{tax2004support}, inspired by Support Vector Machines (SVM), is a classical anomaly detection method. During training, SVDD constructs a hypersphere to encapsulate normal data points. During prediction, the hyperspherical boundary is used to classify test samples. Samples outside the hypersphere, i.e., samples whose distance to the center exceeds the radius, are identified as outliers, while those inside are classified as normal. Using kernel functions, SVDD effectively captures non-linear relationships within the data.

Deep SVDD \cite{ruff2018deep} extends shallow SVDD to a deep learning framework, achieving better data representation and classification results. The Deep SVDD training framework is illustrated in Fig. \ref{DeepSVDDfig}. The input space consists solely of normal data, which is transformed by a neural network such that the output space data is enclosed within a hypersphere.
    
Given the input space $\mathcal{X} \subseteq \mathbb{R}^d$ and the output space $\mathcal{F} \subseteq \mathbb{R}^p$, $\phi(\cdot;\mathcal{W}):\mathcal{X} \rightarrow \mathcal{F}$ denotes the mapping of the neural network with hidden layers $L\in \mathbb{N}$ from the space $\mathcal{X}$ to the space $\mathcal{F}$ and $\mathcal{W}=\{\mathbf{W}^1, \cdots,\mathbf{W}^L\}$. $\mathbf{W}^l$ is the weight matrix of the $l$-th layer. 
 
Given the dataset $\mathcal{D}=\{\mathbf{x}_{1}, \ldots, \mathbf{x}_{n}\}$, Deep SVDD is formulated as the following optimization problem.
\begin{equation}
\label{eq2}
\begin{aligned}
\min_{R, \mathbf{c}, \mathcal{W}} & R^{2}+\frac{1}{v n} \sum_{i=1}^{n} \max\{0,\|\phi(\mathbf{x}_{i} ; \mathcal{W})-\mathbf{c}\|^{2}-R^{2}\} \\
& +\frac{\lambda}{2} \sum_{\ell=1}^{L}\|\mathbf{W}^{\ell}\|_{F}^{2},
\end{aligned}
\end{equation}
where $R$ and $\mathbf{c}$ denote the radius and center of the hypersphere, respectively. The hyperparameter $v\in(0,1]$ controls the proportion of outliers in the model. The objective function consists of three terms: (1) minimizing the volume of the hypersphere, (2) penalizing normal data points mapped outside the sphere, where a penalty is applied if $\|\phi(\mathbf{x})-\mathbf{c}\|>R$, and (3) a weight decay regularizer on the network parameters $\mathcal{W}$ with hyperparameter $\lambda >0$. 

For anomaly detection, once the model is trained with optimal parameters \(\mathcal{W}^{*}\), \(\mathbf{c}^*\), and \(R^*\), the anomaly score for a test point \(\mathbf{x}\) is defined as:
\begin{equation}\label{Decision}
s(\mathbf{x}) = \|\phi(\mathbf{x}; \mathcal{W}^{*}) - \mathbf{c}^*\|^{2} - R^{*2}.
\end{equation}
A test point \(\mathbf{x}\) with \(s(\mathbf{x}) > 0\) is identified as an anomaly, as it lies outside the hypersphere. Conversely, points with \(s(\mathbf{x}) \leq 0\) are classified as normal.

\subsection{Deficiencies of the Vanilla Deep SVDD}
In this subsection, we analyze the problems of the vanilla Deep SVDD.

The first issue is that Deep SVDD is prone to hypersphere collapse. The optimization problem in (\ref{eq2}) treats all samples $\mathbf{x}_i$ as belonging to the normal class. This is the fundamental assumption of the one-class model. However, due to the powerful feature extraction capability of neural networks, there is a risk that all samples will be mapped to the same point. As a result, the term $\|\phi(\mathbf{x}; \mathcal{W}^{*}) - \mathbf{c}^*\|$ in the decision function (\ref{Decision}) becomes constant, and $R^{*}$ reduces to zero. Consequently, the anomaly detector fails to perform correct classification.

The second issue lies in the insufficient precision of the solution derived from Deep SVDD. In both Deep SVDD and its variants, the parameters $R$ and $\mathbf{c}$ are not treated as neural network parameters. Instead, they are obtained through heuristic estimation. Specifically, $\mathbf{c}$ is computed as the average of sample representations after passing through an initialized neural network, while $R$ is set as a corresponding quantile value. Such heuristic estimates do not constitute an optimal solution to the model. From the perspective of the decision function (\ref{Decision}), the final prediction outcome depends heavily on $\mathbf{c}$ and $R$. Consequently, this relatively imprecise heuristic approach significantly impairs prediction accuracy. This situation is illustrated in Fig. \ref{DeepSVDDfig}. The blue and black lines represent the ground truth hypersphere and the solution estimated by Deep SVDD, respectively. In contrast to the ground truth, the Deep SVDD result exhibits a pronounced deviation. 

The third limitation is that the Deep SVDD model operates as a black box, and there is a notable lack of interpretability for both the model's internal mechanisms and its output decisions.

\section{Proposed Method}
\label{ref3}
In this paper, we address the aforementioned limitations of Deep SVDD by introducing three key improvements: (1) leveraging a small set of readily available abnormal samples to refine the decision boundary through the application of the maximum margin principle; (2) developing an end-to-end optimization algorithm to obtain more accurate solutions; and (3) demonstrating the intrinsic interpretability of our model, thereby enhancing its transparency.

\subsection{Implementation of Maximum Margin Principle}
We use $\phi(\cdot;\mathcal{W}):\mathcal{X} \rightarrow \mathcal{F}$ to denote the mapping of the neural network from space $\mathcal{X}$ to space $\mathcal{F}$. Given a data set $\mathcal{D}=\{\mathbf{x}_{1}, \ldots, \mathbf{x}_{n}\}$, containing normal data $\{\mathbf{x}_{1}, \ldots, \mathbf{x}_{m}\}$ and abnormal data $\{\mathbf{x}_{m+1}, \ldots, \mathbf{x}_{n}\}$, $n-m \ll m$. IMD-AD introduces the maximum margin principle by adding a margin parameter \(\rho\), to characterize the separation between normal and anomalous samples. The schematic diagram of our model is shown in Fig. \ref{ADMASfig}. Compared to the diagram in Fig. \ref{DeepSVDDfig}, there are mainly two differences. Firstly, a small number of obtainable abnormal training samples are added, making the decision boundary more accurate. At the same time, through the design of the maximum margin, the collapse of the hypersphere could be avoided.

The IMD-AD is formulated as the following optimization problem:
\begin{equation}\label{eq7}
\begin{aligned}
&\underset{\overline{R},\mathbf{c},\overline{\rho},\mathcal{W}}{\text{min}}~~~\overline{R}-v\overline{\rho}+\frac{1}{v_1m}\sum_{i=1}^{m}\text{max}\{0,\|\phi(\mathbf{x}_i; \mathcal{W})-\mathbf{c}\|^2-\overline{R}\}\\
&+\frac{1}{v_2(n-m)}\sum_{i=m+1}^{n}\text{max}\{0,\overline{R}+\overline{\rho}-\|\phi(\mathbf{x}_i; \mathcal{W})-\mathbf{c}\|^2\}\\
&+\frac{\lambda}{2}\sum_{\ell=1}^{L}\|\mathbf{W}^{\ell}\|_F^2\\
&\text{s.t.}~~~\overline{R}>0,~\overline{\rho}>0.
\end{aligned}
\end{equation}
In Eq. (\ref{eq7}), $\phi(\mathbf{x}_i; \mathcal{W})$ denotes the feature mapping of the training sample $\mathbf{x}_i$ obtained through the neural network. Let $d_i = \|\phi(\mathbf{x}_i; \mathcal{W}) - \mathbf{c}\|$ represent the distance between the mapped feature and the center $\mathbf{c}$ of the hypersphere. As illustrated in Fig. \ref{ADMASfig}, normal samples are expected to lie within a hypersphere of radius $R$, i.e., $d_i^2 \leq R^2$. To enforce separation between normal and anomalous samples, we introduce a margin parameter $\rho$ and construct an enlarged concentric hypersphere centered at $\mathbf{c}$ with radius $\sqrt{R^2 + \rho^2}$. Anomalous samples are required to satisfy $d_i^2 \geq R^2 + \rho^2$, ensuring that they fall outside this larger hypersphere. Maximizing $\rho^2$ increases the separation between the two classes, enhancing the model's discriminative power.

For computational convenience, we define $\overline{R} = R^2$ and $\overline{\rho} = \rho^2$. The loss function for a normal sample is defined as $\max \{0, d_i^2 - \overline{R} \}$, which penalizes those lying outside the smaller hypersphere. For an anomalous sample, the loss is given by $\max \{0, \overline{R} + \overline{\rho} - d_i^2 \}$, penalizing those located inside the extended boundary.

In the overall objective function, minimizing $\overline{R}$ encourages normal samples to cluster tightly around the center, while maximizing $\overline{\rho}$ (or equivalently minimizing $-\overline{\rho}$) enforces a clear separation from anomalies. To prevent overfitting, a regularization term $\frac{\lambda}{2} \sum_{\ell=1}^L \|\mathbf{W}^{\ell}\|_F^2$ is introduced to constrain model complexity. The hyperparameters $v, v_1, v_2$, and $\lambda$ are all tunable and lie in the interval $(0, 1]$. The properties of these parameters will be proved in Section \ref{vSection}. 

Once the model is trained with optimal parameters \(\mathcal{W}^{*}\), \(\mathbf{c}^*\), \(\overline{R}^*\), and \(\overline{\rho}^*\), the anomaly score for a test point \(\mathbf{x}\) is defined as:
\begin{equation}\label{eq8}	 
s(\mathbf{x})=\|\phi(\mathbf{x};\mathcal{W}^*)-\mathbf{c}^*\|^2-T^2,
\end{equation}
where $T=\overline{R}^*+(\sqrt{\overline{R}^* + \overline{\rho}^*} - \sqrt{\overline{R}^*})/2$.

Compared with Eq. (\ref{Decision}), the decision boundary from our function (\ref{eq8}) is no longer a hypersphere with radius $R$, but with radius $T$. As shown in Fig. \ref{ADMASfig}, the decision boundary is located in the middle of a small sphere containing normal samples and a large sphere excluding abnormal samples. The advantage of this approach is that even if the small sphere undergoes hypersphere collapse (i.e. $R=0$), $T$ will not be zero, allowing our method to tolerate the collapse problem.
% 相比 Eq. (\ref{Decision}), 决策函数(\ref{eq8})的分类边界不再是以$R$为半径的超球, 而是以$T$为半径。如图Fig. \ref{ADMASfig} 所示, 决策边界是在包含正常样本的小球和将异常样本排除在外的大球的中央。这样做的好处在于，即使小球发生了超球崩塌，即$R=0$, 这时候$T$也不会为0，使得我们的方法可以容忍崩塌问题。
\subsection{Efficient End-to-end Optimization}
\subsubsection{Neural Network Parameterization of the Hypersphere}
In optimization problem (\ref{eq7}), the variables to be optimized can be categorized into two main components: the neural network weights $\mathcal{W}$ responsible for feature extraction, and the hypersphere parameters $\overline{R}, \mathbf{c}, \overline{\rho}$ that define the decision boundary. Deep SVDD treats the feature extraction and hypersphere construction as separate stages, which weakens the overall cohesiveness of the model and limits the interpretability of its parameters. Moreover, its solution procedure relies on alternating optimization, where the network parameters and hypersphere parameters are solved independently, often using heuristics such as the mean and quantiles of the feature distribution, which can lead to suboptimal classification boundaries and reduced training efficiency.

To address these limitations, we conduct a more detailed analysis of the hypersphere-based decision component in model (\ref{eq7}), aiming to integrate it more tightly with the neural network feature extraction process. This joint formulation not only enhances the interpretability of the model but also lays a solid foundation for developing an effective end-to-end training algorithm.

First, the squared distance between the extracted feature $\phi(\mathbf{x}; \mathcal{W})$ with respect to the hypersphere boundary can be represented as 
\begin{equation}\label{eq9}
\begin{aligned}
&\|\phi(\mathbf{x}; \mathcal{W})-\mathbf{c}\|^2-\overline{R}\\
&=\mathbf{c}^\top \mathbf{c}-2\mathbf{c}^\top \phi(\mathbf{x};\mathcal{W})-\overline{R}+\|\phi(\mathbf{x};\mathcal{W})\|^2.
\end{aligned}
\end{equation}

To establish a relationship between the hypersphere parameters $(\mathbf{c}, R)$ and the final layer parameters of the neural network, we define
\begin{equation}\label{eq11}
\mathbf{w}=-2\mathbf{c}, ~~b=\mathbf{c}^\top \mathbf{c}-\overline{R}.
\end{equation}

According to Eq. (\ref{eq11}), $\mathbf{w}$ and $b$ are inherently coupled, i.e., $b = \frac{1}{4} \mathbf{w}^\top \mathbf{w} - \overline{R}$. However, during neural network training, it is preferable to decouple these parameters. Additionally, in practice, it is important to normalize the weights and bias of the last layer to avoid gradient explosion, which can hinder convergence. To this end, we normalize the hypersphere center by enforcing $\mathbf{c}^\top \mathbf{c} = 1$, which yields
\begin{equation}
\mathbf{c}=-\frac{\mathbf{w}}{2},~~ \overline{R}=1-b.
\end{equation}

Now define the output of the neural network as
\begin{equation}\label{gx}
\begin{aligned}
g(\phi(\mathbf{x}; \mathcal{W});\mathbf{w},b)&= \mathbf{c}^\top \mathbf{c}-2\mathbf{c}^\top \phi(\mathbf{x};\mathcal{W})-\overline{R} \\
&= \mathbf{w}^\top \phi(\mathbf{x};\mathcal{W})+b.
\end{aligned}
\end{equation}

Here, $\phi(\mathbf{x}; \mathcal{W})$ and $g(\phi(\mathbf{x}; \mathcal{W}); \mathbf{w}, b)$ are regarded as the outputs of the penultimate and final layers of the neural network, respectively. Therefore, Eq. (\ref{eq9}) can be rewritten as
\begin{equation}\label{eq10}
\begin{aligned}
&\|\phi(\mathbf{x}; \mathcal{W})-\mathbf{c}\|^2-\overline{R}\\
&=g(\phi(\mathbf{x}; \mathcal{W});\mathbf{w},b)+\|\phi(\mathbf{x};\mathcal{W})\|^2.
\end{aligned}
\end{equation}

With this formulation, we effectively embed the hypersphere parameters into the neural network, establishing a direct connection between the decision boundary and the network parameters.

Substituting the above definitions into Eq. (\ref{eq7}), the optimization objective becomes:
\begin{equation}\label{eq12}
\begin{aligned}
&\underset{\mathcal{W},\mathbf{w},b,\overline{\rho}}{\text{min}}~~~\frac{1}{v_1 m}\sum_{i=1}^{m}\text{max}\{0,\|\phi(\mathbf{x}_i;\mathcal{W})\|^2+g(\mathbf{x}_i)\}\\
&+\frac{1}{v_2(n-m)}\sum_{i=m+1}^{n}\text{max}\{0,\overline{\rho}-\|\phi(\mathbf{x}_i;\mathcal{W})\|^2-g(\mathbf{x}_i)\}\\
&+1-b-v\overline{\rho}+\frac{\lambda}{2}\sum_{l=1}^{L}\|\mathbf{W}^l\|_F^2\\
&s.t.~~~\mathbf{w}^{\top}\mathbf{w}=4,~b<1,~\overline{\rho}>0,
\end{aligned}
\end{equation}
where $\mathbf{w}$ and $b$ represent the weight and bias of the last neural network layer, respectively. The first term penalizes normal samples that fall outside the hypersphere, while the second penalizes anomalous samples that fall inside the extended margin boundary. The third term minimizes the hypersphere volume by maximizing $b$ (subject to $b < 1$), while simultaneously maximizing $\overline{\rho}$ to increase the margin between anomalies and the hypersphere. The final term regularizes the neural network weights to prevent overfitting. The constraints follow from the normalization condition $\mathbf{c}^\top \mathbf{c} = 1$, and ensure the radius and margin remain positive.

\subsubsection{End-to-end Training of IMD-AD}
Compared to standard unconstrained deep learning models, the optimization problem in Eq. (\ref{eq12}) includes explicit constraints, which prevents the direct application of conventional backpropagation-based gradient descent. To address this, we adopt the Lagrange multiplier method to reformulate the constrained problem into the following unconstrained optimization problem:

\begin{equation}
\max_{\alpha,\beta,\gamma} \inf_{\mathcal{W}, \mathbf{w}, b, \overline{\rho}} \mathcal{L}_2(\mathbf{x}; \mathcal{W},\mathbf{w},b,\overline{\rho}; \alpha, \beta, \gamma)
\end{equation}

where
\begin{equation}\label{eq13}
\begin{aligned}
&\mathcal{L}_2(\mathbf{x}; \mathcal{W},\mathbf{w},b,\overline{\rho}; \alpha, \beta, \gamma)\\
&=1-b-v\overline{\rho}+\frac{1}{v_1m}\sum_{i=1}^{m}\text{max}\{0,\|\phi(\mathbf{x}_i;\mathcal{W})\|^2+g(\mathbf{x}_i)\}\\
&+\frac{1}{v_2(n-m)}\sum_{i=m+1}^{n}\text{max}\{0,\overline{\rho}-\|\phi(\mathbf{x}_i;\mathcal{W})\|^2-g(\mathbf{x}_i)\}\\
&+\frac{\lambda}{2}\sum_{l=1}^{L}\|\mathbf{W}^l\|_F^2+\alpha(\mathbf{w}^{\top}\mathbf{w}-4)+\beta(b-1)-\gamma\overline{\rho}.
\end{aligned}
\end{equation}
Here, $\alpha\ne 0,~\beta\ge 0,~\gamma\ge 0$ are Lagrange multipliers. 

We use $\boldsymbol{\theta} = \{\mathcal{W}, \mathbf{w}, b, \overline{\rho}\}$ to denote all trainable parameters of the neural network. In the proposed algorithm, the Lagrangian function $\mathcal{L}_2(\boldsymbol{\theta})$ is minimized via backpropagation. Simultaneously, the dual variables $\alpha$, $\beta$, and $\gamma$ are updated by maximizing $\mathcal{L}_2(\alpha, \beta, \gamma)$ using the projected gradient ascent method. The gradients of the Lagrangian function $\mathcal{L}_2$ with respect to $\alpha$, $\beta$, and $\gamma$ are given by:
\begin{equation}\label{eq14}
\begin{aligned}
&\frac{\partial \mathcal{L}_2}{\partial \alpha}=\mathbf{w}^{\top}\mathbf{w}-4,\\
&\frac{\partial \mathcal{L}_2}{\partial \beta}=b-1,\\
&\frac{\partial \mathcal{L}_2}{\partial \gamma}=-\overline{\rho}.
\end{aligned}
\end{equation}

The pseudo-code of IMD-AD is summarized in Algorithm 1. It's important to note that in practice, we also set the parameter $\overline{\rho}$ as a learnable neural network parameter, which is initialized alongside the neural network. 
%For a given new test point $\mathbf{x}\in \mathbb{R}^d$, we determine whether it is an outlier by calculating the anomaly score using Eq. (\ref{eq8}).

\begin{algorithm}[t]
\caption{Training process of IMD-AD}
\begin{algorithmic}[1]
\REQUIRE Normal training data $D_{1}$, abnormal training data $D_{2}$, validation data $D_{3}$
\ENSURE Neural network parameters $\boldsymbol{\theta}$
\HYPERPARAMS  $\lambda \geq 0$, $\nu$, $\nu_{1}$, $\nu_{2} \geq 0$, epochs $N$, batch size $M$
\STATE Randomly initialize neural network weights $\boldsymbol{\theta}$
\STATE Initialize learning rate $r$, a fixed epochs $k$, Lagrange multipliers $\alpha$, $\beta$, and $\gamma$
\FOR{epoch = 1 to $N$}
    \FOR{batch = 1 to $M$}
        \STATE \textbf{Forward Pass:}
        \STATE Compute $\phi(x), g(x)=\texttt{net}(x)$               
        \STATE $\boldsymbol{\theta} = \texttt{get\_parameter(net)}$
        \STATE Compute training loss: $\mathcal{L}_{2}(\boldsymbol{\theta})$ of Eq. (\ref{eq13})
        \STATE \textbf{Backward Pass:}
        \STATE Compute $\nabla_{\boldsymbol{\theta}} \mathcal{L}_{2}(\boldsymbol{\theta})$
        \STATE Update parameters $\boldsymbol{\theta}$ using Adam optimizer with learning rate $r$
    \ENDFOR
    \IF{epoch mod $k$ == 0}
        \STATE Update Lagrange multipliers:
        \STATE $\alpha_{t+1} = \alpha_t + r(\mathbf{w}^{\top}\mathbf{w} - 4)$
        \STATE $\beta_{t+1} = [\beta_t + r(b - 1)]_+$
        \STATE $\gamma_{t+1} = [\gamma_t - r\overline{\rho}]_+$
    \ENDIF
    \IF{the loss $\mathcal{L}_{2}$ on $D_3$ stops decreasing for $k$ consecutive epochs}
        \STATE Early stop
    \ENDIF
\ENDFOR
\end{algorithmic}
\end{algorithm}

\subsection{Interpretability of IMD-AD}
The interpretability is discussed in further detail in terms of model architecture and model hyperparameters.

\subsubsection{Intrinsic Interpretability of the IMD-AD Architecture}
Most existing deep anomaly detection methods are regarded as black-box models, whose internal mechanisms are difficult to interpret. Recent studies on the interpretability of deep learning models can generally be categorized into post hoc explanations and intrinsic interpretability \cite{zhang2021survey}. Our proposed IMD-AD inherently incorporates intrinsic interpretability. It embraces model decomposability, which refers to the ability to interpret each component of the AD layer. Specifically, the input, weight parameters, and loss function are all interpretable: the input corresponds directly to the given data; the weights $\mathbf{W}^l$ ($l=1,\cdots,L-1$) serve for feature extraction; and the weight $\mathbf{W}^L$ in the final layer can be interpreted as the center and radius of the hypersphere used for anomaly detection. The loss function is reformulated from the traditional anomaly detection method SVDD and implements the maximum margin principle. The interpretability verification experiment is presented in Section \ref{InterAnaly}. As illustrated in Figs. \ref{fig3}, \ref{fig:mnist_train_epoch}, and \ref{fig:mnist_results_KDE}, the reconstruction of the hypersphere center, the boundaries of the normal and abnormal classes, and the decision boundary are clearly depicted, demonstrating that our method effectively captures intrinsic sample characteristics and achieves good separation between the two classes.

Furthermore, IMD-AD exhibits algorithmic transparency, as its error surface and dynamic behavior can be mathematically reasoned about, enabling users to understand the model's decision process. In contrast to parameters traditionally set heuristically in Vanilla Deep SVDD, our model parameters can be optimally determined through efficient end-to-end joint optimization. Numerical verification of training convergence and result stability is provided in Fig. \ref{fig4}.

%We can also understand how IMD-AD works from the perspective of the popular attention mechanism \cite{NEURIPS2023_970f59b2}. IMD-AD aims to learn a hyperplane determined by the center $\mathbf{c}$ of the normal sample and the anomaly detection boundary with parameters $\mathbf{W}^L$. The equivalence relation between hypersphere and hyperplane in SVDD is discussed in \cite{taxSupportVectorData2004}. Hyperplanes are able to group data points of normal classes together and exclude abnormal data points based on attention. More specifically, to learn the hyperplane through the optimization problem (\ref{eq12}), IMD-AD computes the distance between the input $\mathbf{x}_i$ and the center of the normal class $\mathbf{c}$ via Eq. (\ref{gx}). After that, the loss of IMD-AD is constructed based on the anomaly score values of all samples. Actually, the attention here serves as the anomaly detection assignment. 

\subsubsection{$\nu$-Property of IMD-AD}\label{vSection}
\label{ref4}
Here, we present a theoretical analysis to clarify the role and interpretability of $\nu$, $\nu_1$ and $\nu_2$. They appear as the tunable parameters in Eq. (\ref{eq13}).

Define $n_{out}^{+}$ as the number of normal class samples located outside the boundary of the small hypersphere with radius $R$. Define $n_{out}^{-}$ as the number of anomalous samples located within the boundary of the large hypersphere with radius $\sqrt{R^{2}+\rho^{2}}$. Then, the following property holds:

\textbf{Theorem ($\nu$-Property).} For IMD-AD, $\left(\nu + 1\right) \nu_1$ is an upper bound on the fraction of normal samples being outside the small hypersphere. $\nu \nu_2$ is an upper bound on the fraction of abnormal samples being in the large hypersphere. That is,
\begin{equation}
\begin{gathered}
\frac{n_{out}^{+}}{m} < \left(\nu + 1\right) \nu_1, \\
\frac{n_{out}^{-}}{n-m} < \nu \nu_2,
\end{gathered}
\label{eq:nu_property}
\end{equation}
where $\nu \in [0, +\infty)$, $\nu_1 \in (0, \frac{1}{\nu + 1}]$, $\nu_2 \in (0, \frac{1}{\nu}]$.

\textbf{Proof.} The objective function of (\ref{eq7}) can be transformed as follows,
\begin{equation}
\begin{aligned}
&J(\bar{R}, \overline{\rho}) = \left(1-\frac{n_{out}^{+}}{\nu_1 m}+\frac{n_{out}^{-}}{\nu_2 (n-m)}\right) \bar{R} + \sum^{n_{out}^{+}}_{i}d_i - \sum^{n_{out}^{-}}_{j}d_j\\
&+ \left(\frac{n_{out}^{-}}{\nu_2(n-m)}-\nu\right) \overline{\rho} 
\end{aligned}
\label{eq:transformed_obj}
\end{equation}
where $d_i = \|\phi(\mathbf{x}_i; \mathcal{W}) - \mathbf{c}\|$ represents the distance from the feature-mapped sample to the center of the hypersphere.

To construct a compact hypersphere that encompasses the normal class while maximizing the margin between the anomalous and normal classes, we aim to minimize Eq. (\ref{eq:transformed_obj}). To ensure a meaningful solution, the following conditions must hold:
\begin{equation}
1 - \frac{n_{out}^{+}}{\nu_1 m} + \frac{n_{out}^{-}}{\nu_2 (n - m)} > 0,
\label{eq:condition1}
\end{equation}
\begin{equation} 
\frac{n_{out}^{-}}{\nu_2(n - m)} - \nu < 0.
\label{eq:condition2}
\end{equation}

After a straightforward transformation, Eq. (\ref{eq:nu_property}) is obtained.

Additionally, since $0 < \left(\nu + 1\right)\nu_1 \leq 1$, $0 <\nu\nu_2 \leq 1$, we have $0 < \nu_1 \leq 1/(\nu + 1)$ and $0 <\nu_2 \leq 1/\nu$.

% \section{Discussion}
% \label{ref5}
% While Deep SVDA also leverages abnormal samples to refine the classification boundaries, IMD-AD goes a step further by maximizing the margin between the abnormal samples and the boundaries. This intentionally introduces a certain degree of tolerance for errors in the classification boundary. Therefore, when dealing with the same data, the IMD-AD model demonstrates greater resilience, surpassing the performance of Deep SVDA. This claim has been substantiated through the experiments conducted in Section 4.3.

% In previous hypersphere-based deep anomaly detection methods such as Deep SVDD \cite{ruff2018deep} and Deep Semi-supervised Anomaly Detection (Deep SAD) \cite{ruff2019deep}, the radius and center of the hypersphere are difficult to precisely compute. Our algorithm decomposes the anomaly score formula and defines a portion of it as a neural network layer, which is added to the output of the neural network mapping. This allows us to accurately determine the center and radius of the sphere by extracting the network parameters from the final layer's output. The algorithm we propose can significantly enhance the performance of the model, particularly when applied to large-scale high-dimensional datasets (refer to Section 5.2).

% In addition, the neural network constructed through this algorithm possesses unique interpretability. The parameters of the last layer of the neural network can be interpreted as functions of the center and radius of a hypersphere, giving practical meaning to the network parameters.

\section{Numerical Experiments}
\label{ref5}
The experiments are mainly conducted in three parts: prediction performance comparions, interpretability analysis, and ablation experiments.

\subsection{Experimental Design}

\subsubsection{Datasets}
Numerical experiments include three image datasets, i.e., MNIST, Fashion MNIST and CIFAR-10, three tabular datasets, i.e., OBS Network (ON), Cardiotocography (Card), and Breast Cancer (BC), and two synthetic Datasets, i.e., Moon and Spiral datasets. 
\subsubsection{Baselines}

The effectiveness of our proposed IMD-AD is benchmarked against eight existing methods, comprising five shallow and three deep learning-based anomaly detection approaches.

\begin{itemize}
\item Isolation Forest (IF) \cite{liu2008isolation}. It is a tree-based method. The experiments of IF set the parameters $n$-estimators to 100, max-samples to 256, and contamination to 0.1.
\item KDE \cite{latecki2007outlier}. It is a non-parametric statistical method for estimating unknown density functions. The Gaussian kernel is used in KDE, and the tuning of the kernel bandwidth parameters is in $\left\{2^{0.5}, 2^{1}, \ldots, 2^{5}\right\}$.
\item OCSVM \cite{scholkopf1999support}. It is an anomaly detection method based on SVM. The hyperparameter $\nu$ is selected in $\{0.1,0.2,\cdots,0.9\}$ for training.
\item SVDD \cite{tax2004support}. This is another type of one-class classification method. The optimal parameter $\nu$ of SVDD is selected in $\{0.1,0.2,\cdots,0.9\}$.
\item SSLM \cite{wu2009small}. This method involves incorporating a small number of abnormal samples in the training process and is considered a shallow approach. The best parameters $\nu$, $\nu_1$, and $\nu_2$ of SSLM are selected in $\{0.1,0.2,\cdots,0.9\}$.
\item Autoencoder (AE) \cite{an2015variational}. The idea of AE for anomaly detection is to use reconstructed mean square error as the anomaly score, assuming that abnormal samples cannot be reconstructed well.
\item Deep SVDD \cite{ruff2018deep}. The selection of the parameter $\nu$ in the range of $\{0.1,0.2,\cdots,0.9\}$. To prevent the collapse of the sphere, the value of the center $\mathbf{c}$ of the hypersphere is set to the average value of the mapped data after performing an initial forward pass, and the radius $R$ is estimated by quantile of extracted normal samples.
\item DROCC \cite{goyal2020drocc}. This is a robust one-class classification method that addresses the hypersphere collapse problem through adversarial training. We select the radius $r$ proportional to $\sqrt{d}/2$, where $d$ is the input dimension, to approximate the expected inter-sample distance under normalized features.  
The projection factor $\gamma$ is set to $2$ unless otherwise noted.  
The perturbation scale $\lambda$ is chosen from $\{0.5, 1.0\}$.  
Gradient ascent uses step sizes in $\{0.1, 0.01\}$, while descent steps are drawn from $\{10^{-2}, 10^{-4}\}$. We tune the optimizer in {Adam, SGD}.
\end{itemize}

\subsubsection{Implementation Details}
For each dataset, we designate one class as the normal class and treat all remaining classes as abnormal. During training, we randomly sample 10\% of the abnormal data for model training. The test set includes samples from all classes. All deep learning models adopt the same network architecture. For image datasets, a convolutional neural network (CNN) is used for feature mapping, while for tabular and synthetic datasets, a fully connected (FC) neural network is employed. Detailed dataset and architecture information is provided in Table~\ref{data}. For the first three datasets, we employed the default training-test split strategy provided. For the remaining tabular datasets and synthetic datasets, we performed five-fold cross-validation. Parameter selection with grid search. For IMD-AD, the Adam optimizer is used with an initial learning rate of 0.0001. The learning rate is dynamically adjusted using the MultiStepLR scheduler. The maximum number of training epochs is set to 200. The batch size is 50 for tabular and synthetic datasets, and 150 for image datasets. The weight decay hyperparameter is set to $\lambda = 5 \times 10^{-6}$. Hyperparameters $\nu$, $\nu_1$, and $\nu_2$ are selected based on the $\nu$-Property described in Section~\ref{vSection}, with values chosen from the range $\{0.1, 0.2, \ldots, 0.9\}$.  

\begin{table}[ht]
\setlength{\tabcolsep}{3pt} % 默认是6pt，减小这个值

\centering
\caption{Overview of Datasets and Corresponding Network Architecture.}
\scalebox{0.85}{ % 缩放因子
\begin{tabular}{lrrrr}
\hline
Datasets & \#Classes & Sample size & Dimensions & Hidden layers \\
\hline
MNINST& 10 & 60000 (train) + 10000 (test) & 28*28 & CNN$\times$2+FC$\times$3 \\
Fashion MNIST & 10 & 60000 (train) + 10000 (test) & 28*28 & CNN$\times$2+FC$\times$3 \\
CIFAR-10 & 10 & 50000 (train) + 10000 (test) & 32*32 & CNN$\times$3+FC$\times$3 \\
ON& 4 & 1060 & 21 & FC$\times$3 \\
Card& 10 & 2126 & 21 & FC$\times$3 \\
BC& 2 & 569 & 30 & FC$\times$3 \\
%CC& 2 & 1000 & 20 & FC$\times$3 \\
Two Moon& 2 & 1000 & 2 & FC$\times$3 \\
Spiral& 2 & 1000 & 2 & FC$\times$3 \\
\hline
\end{tabular}%
}
\label{data}%
\end{table}

\subsection{Prediction Performance Comparisons}
The experimental results of Area Under the receiver operating characteristic Curve (AUC) are presented in Tables~\ref{tab_all} and \ref{tab4}.

Table~\ref{tab_all} presents the prediction performance of nine methods on three benchmark image datasets. It can be observed that our proposed IMD-AD consistently ranks within the top two positions, whereas the rankings of other methods are unstable. On the Fashion MNIST dataset, our method outperforms the second-best OCSVM by an average AUC improvement of 3.93\%. On CIFAR-10, our method surpasses the second-ranked DROCC by an average AUC margin of 9.62\%. Although DROCC achieves a slightly higher average AUC on MNIST, the margin is only 0.28\%. Moreover, our IMD-AD attains an AUC above 99.2\% across all classes in this dataset. These results demonstrate that the proposed method achieves high predictive performance and exhibits greater stability across diverse scenarios. DROCC takes second place in terms of average ranking. The shallow OCSVM ranks third, while Deep SVDD and KDE methods tie for fourth place. The shallow SVDD ranks last. This indicates that our improvements to Deep SVDD, i.e., incorporating the maximum margin principle and precise parameter optimization, are effective and significantly enhance the predictive performance of the original Deep SVDD.

\begin{table}[!t]
	\setlength{\tabcolsep}{2pt}
	\caption{AUC (\%) OF NINE METHOD ON THREE IMAGE DATASETS.}
	\label{tab_all}
	\centering
	\scalebox{0.8}{
		\begin{tabular}{@{}l|ccccccccc@{}}
			\hline
			\textbf{  } & \multicolumn{9}{c@{}}{\textbf{Methods}} \\
			\cline{2-10}
			\textbf{MNIST} & IF & KDE & OCSVM & SVDD & SSLM & AE & Deep SVDD & DROCC & IMD-AD \\
			\hline
			0 & 90.33 & 96.28 & 99.01 & 90.39 & 93.88 & 97.47 & 98.00  & \textbf{100.00} & 99.86 \\
			1 & 97.46 & 92.54 & 99.56 & 98.34 & 98.61 & 96.93 & 99.70  & \textbf{99.96} & 99.93 \\
			2 & 78.96 & 85.56 & 95.78 & 53.33 & 85.49 & 80.06 & 91.70  & \textbf{100.00} & 99.66 \\
			3 & 82.12 & 87.17 & 93.96 & 72.44 & 83.67 & 86.94 & 91.90  & \textbf{100.00} & 99.66 \\
			4 & 80.35 & 82.34 & 95.79 & 84.50 & 89.04 & 83.73 & 94.90  & \textbf{100.00} & 99.90 \\
			5 & 74.24 & 73.05 & 91.57 & 59.91 & 74.10 & 79.22 & 88.50  & \textbf{100.00} & 99.67 \\
			6 & 84.92 & 91.27 & 98.09 & 70.66 & 90.70 & 94.96 & 98.30  & \textbf{100.00} & 99.80 \\
			7 & 85.57 & 86.66 & 96.01 & 70.67 & 87.76 & 90.61 & 94.60  & \textbf{100.00} & 99.69 \\
			8 & 81.43 & 89.16 & 93.76 & 78.56 & 85.06 & 86.51 & 93.90  & \textbf{100.00} & 99.73 \\
			9 & 84.15 & 88.59 & 96.34 & 81.57 & 85.32 & 93.54 & 96.50  & \textbf{100.00} & 99.29 \\
			\cdashline{1-10}
			\textit{Avg.}& 83.95 & 87.26 & 95.99 & 76.04 & 87.36 & 89.00 & 94.80 & \textbf{100.00} & 99.72 \\
			\cdashline{1-10}
			\textit{Rank} & 8 & 7 & 3 & 9 & 6 & 5 & 4 & 1 & 2 \\
			\hline
			
			\textbf{Fashion MNIST} & \multicolumn{9}{c}{} \\
			\hline
			T-shirt     & 91.25 & 91.57 & 91.87 & 82.81 & 86.71 & 88.70 & 87.46 & 81.41 & \textbf{96.37} \\
			Trouser     & 97.75 & 98.90 & 99.02 & 83.00 & 96.03 & 96.58 & 96.98 & 85.74 & \textbf{99.62} \\
			Pullover    & 87.33 & 88.82 & 89.16 & 86.09 & 84.99 & 85.81 & 84.86 & 87.02 & \textbf{93.56} \\
			Dress       & 93.55 & 93.94 & 94.01 & 80.66 & 92.05 & 88.55 & 90.16 & 90.47 & \textbf{97.13} \\
			Coat        & 89.92 & 90.08 & 90.68 & 85.94 & 86.60 & 84.13 & 87.07 & 80.82 & \textbf{94.74} \\
			Sandal      & 92.58 & 90.70 & 91.78 & 78.26 & 84.20 & 88.64 & 89.60 & 81.95 & \textbf{98.87} \\
			Shirt       & 79.56 & 82.40 & 82.08 & 80.26 & 77.07 & 80.65 & 79.51 & 75.79 & \textbf{86.68} \\
			Sneaker     & 98.27 & 98.60 & 98.68 & 96.00 & 98.17 & 96.34 & 97.52 & 93.12 & \textbf{99.01} \\
			Bag         & 87.28 & 88.58 & 89.87 & 64.67 & 67.64 & 88.71 & 92.04 & 75.77 & \textbf{98.99} \\
			Ankle boot  & 98.06 & 96.79 & 98.03 & 95.86 & 88.72 & 92.53 & 97.45 & 90.62 & \textbf{99.54} \\
			\cdashline{1-10}
			\textit{Avg.}       & 91.56 & 92.04 & 92.52 & 83.36 & 86.22 & 89.06 & 90.27 & 84.27 & \textbf{96.45} \\
			\cdashline{1-10}
			\textit{Rank}        & 4 & 3 & 2 & 9 & 7 & 6 & 5 & 8 & 1 \\
			\hline

			\textbf{CIFAR-10} & \multicolumn{9}{c}{} \\
			\hline
			Airplane   & 75.26 & 64.03 & 67.50 & 62.47 & 77.58 & 63.93 & 61.70 & 82.27 & \textbf{85.88} \\
			Automobile & 58.87 & 61.83 & 62.23 & 42.22 & 61.42 & 60.58 & 65.90 & 72.02 & \textbf{89.34} \\
			Bird       & 60.17 & 49.68 & 51.61 & 64.98 & 68.56 & 51.19 & 50.80 & 68.65 & \textbf{74.27} \\
			Cat        & 55.10 & 58.58 & 56.52 & 49.34 & 63.31 & 54.66 & 59.10 & 68.89 & \textbf{78.55} \\
			Deer       & 63.15 & 63.27 & 59.48 & 74.13 & 76.51 & 55.35 & 60.90 & 74.76 & \textbf{77.21} \\
			Dog        & 65.10 & 64.73 & 59.98 & 50.04 & 65.15 & 58.25 & 65.70 & 64.31 & \textbf{81.66} \\
			Frog       & 69.42 & 70.53 & 63.66 & 69.93 & 77.90 & 62.57 & 67.70 & 76.94 & \textbf{87.12} \\
			Horse      & 64.72 & 63.65 & 64.82 & 52.16 & 64.56 & 63.42 & 67.30 & 72.19 & \textbf{85.40} \\
			Ship       & 79.86 & 77.39 & 80.16 & 66.44 & 81.92 & 79.07 & 75.90 & 80.29 & \textbf{88.69} \\
			Truck      & 72.58 & 74.87 & 74.28 & 49.93 & 71.94 & 70.73 & 73.10 & 77.51 & \textbf{86.00} \\
			\cdashline{1-10}
			\textit{Avg.}      & 66.42 & 64.86 & 64.02 & 58.16 & 70.89 & 61.98 & 64.81 & 73.78 & \textbf{83.40} \\
			\cdashline{1-10}
			\textit{Rank}       & 4 & 5 & 7 & 9 & 3 & 8 & 6 & 2 & 1 \\
			\hline
			\textit{Avg. Rank}  & 5.3 & 5.0 & 4.0 & 9.0 & 5.3 & 6.3 & 5.0 & 3.7 & 1.3 \\
			\hline
		\end{tabular}
	}
\end{table}

Table~\ref{tab4} presents the prediction results of nine methods on three UCI\footnote{https://archive.ics.uci.edu} benchmark tabular datasets. It can be observed that our proposed IMD-AD method consistently ranks within the top two on tabular data, with the second-best being the SSLM method, followed by OCSVM. Our proposed IMD-AD method outperforms the second-ranked SSLM method by an average AUC of $4.82\%$ on the ON dataset and by $3.90\%$ on the Card dataset. Although SSLM achieves slightly better results than IMD-AD on the BC dataset, the margin is only $1.99\%$. Moreover, as shown in Table \ref{tab_all}, SSLM performs substantially lower AUC values on image datasets, ranking only 7th. Additionally, three deep learning methods, i.e., Deep SVDD, DROCC, and AE, perform inadequately on tabular data, whereas our proposed method achieves top-tier performance on both image and tabular datasets. This fully demonstrates the broader applicability of our approach.

\begin{table*}[!h]
	\setlength{\tabcolsep}{4pt}  % 减小列间距（默认是6pt）
	\centering
	\caption{Average AUC (\%) with StdDevs (5 Seeds) per Method on Three Tabular Datasets.}
	\begin{tabular}{llcccccccccccr}
		\hline  
		Datasets & Class &IF &KDE &OCSVM &SVDD &SSLM &AE &Deep SVDD &DROCC &IMD-AD\\
		\hline
		&NB-No Block & 79.89$\pm$3.21 & 79.09$\pm$3.36 & 71.82$\pm$2.36 & 57.39$\pm$3.00 & 81.81$\pm$2.24 & 79.40$\pm$1.94 & 51.26$\pm$7.49 & 74.83$\pm$0.75 & \textbf{88.86$\pm$5.35} \\
		& Block & 88.90$\pm$4.55 & 95.79$\pm$0.89 & 93.49$\pm$1.46 & 92.11$\pm$1.23 & 99.43$\pm$0.33 & 80.46$\pm$1.71 & 76.12$\pm$14.06 & 86.76$\pm$0.69 & \textbf{99.45$\pm$0.53} \\
		ON&No Block & 92.90$\pm$4.42 & 97.86$\pm$0.59 & 96.73$\pm$3.59 & 95.65$\pm$1.83 & \textbf{99.96$\pm$0.05} & 86.74$\pm$4.46 & 84.14$\pm$7.96 & 97.97$\pm$1.64 & 99.92$\pm$0.15 \\
		&NB-Wait & 80.62$\pm$3.48 & 84.90$\pm$3.64 & 83.23$\pm$3.99 & 55.40$\pm$3.59 & 76.77$\pm$5.73 & 83.75$\pm$2.13 & 56.56$\pm$10.99 & 82.36$\pm$0.47 & \textbf{89.00$\pm$2.93} \\
		\cdashline{2-12}
		&\textit{avg.} & 85.58$\pm$6.37 & 89.41$\pm$8.92 & 86.32$\pm$11.25 & 75.14$\pm$21.71 & 89.49$\pm$11.96 & 82.59$\pm$3.33 & 67.02$\pm$15.64 & 85.48$\pm$0.89 & \textbf{94.31$\pm$6.21} \\
		\cdashline{2-12}
		&\textit{Rank} & 5 & 3 & 4& 8 & 2 & 7 & 9 & 6 & 1 \\
		
		\hline
		& 1 & 88.08$\pm$1.53 & 85.90$\pm$1.62 & 85.75$\pm$1.90 & 84.12$\pm$2.91 & 85.98$\pm$1.31 & 86.87$\pm$1.15 & 79.52$\pm$3.73 & 82.94$\pm$1.19 & \textbf{93.19$\pm$2.27} \\
		& 2 & 84.13$\pm$1.46 & 81.82$\pm$2.10 & 87.64$\pm$2.59 & 86.00$\pm$1.30 & 90.32$\pm$1.93 & 78.19$\pm$1.82 & 74.76$\pm$3.33 & 80.98$\pm$0.73 & \textbf{92.19$\pm$1.16} \\
		& 3 & 91.57$\pm$2.27 & 90.83$\pm$3.49 & 90.36$\pm$1.75 & 90.47$\pm$2.79 & 90.27$\pm$3.76 & 82.39$\pm$6.73 & 81.66$\pm$4.43 & 80.51$\pm$0.51 & \textbf{94.87$\pm$1.81} \\
		& 4 & 88.02$\pm$4.44 & 91.98$\pm$4.02 & 93.26$\pm$2.05 & 93.69$\pm$1.64 & 95.61$\pm$1.07 & 79.39$\pm$6.05 & 88.10$\pm$4.41 & 85.92$\pm$0.65 & \textbf{98.37$\pm$0.83} \\
		& 5 & 86.43$\pm$3.89 & 86.79$\pm$1.87 & 87.79$\pm$3.59 & 88.03$\pm$1.87 & 90.26$\pm$1.34 & 86.15$\pm$2.66 & 85.33$\pm$3.24 & 81.11$\pm$1.68 & \textbf{90.73$\pm$3.64} \\
		Card& 6 & 83.83$\pm$3.70 & 81.71$\pm$4.38 & 89.13$\pm$1.97 & 89.28$\pm$1.07 & 90.44$\pm$3.05 & 76.62$\pm$4.22 & 68.57$\pm$1.56 & 85.03$\pm$0.74 & \textbf{92.29$\pm$5.63} \\
		& 7 & 84.36$\pm$3.76 & 84.37$\pm$2.98 & 87.46$\pm$0.82 & 87.52$\pm$3.04 & 93.86$\pm$1.67 & 80.72$\pm$3.14 & 79.94$\pm$3.78 & 89.61$\pm$0.84 & \textbf{98.21$\pm$1.56} \\
		& 8 & 98.06$\pm$0.96 & 97.24$\pm$2.90 & 95.58$\pm$2.51 & 96.57$\pm$1.49 & \textbf{99.57$\pm$0.17} & 52.49$\pm$6.05 & 96.09$\pm$1.96 & 86.94$\pm$0.81 & 99.41$\pm$0.52 \\
		& 9 & 93.80$\pm$2.61 & 94.93$\pm$1.76 & 89.72$\pm$5.77 & 90.52$\pm$3.96 & 93.85$\pm$2.52 & 74.61$\pm$4.21 & 94.71$\pm$2.02 & 91.56$\pm$0.51 & \textbf{98.06$\pm$1.49} \\
		& 10 & 93.17$\pm$2.43 & 92.11$\pm$1.70 & 91.57$\pm$1.71 & 91.42$\pm$1.70 & 92.45$\pm$2.02 & 91.31$\pm$1.90 & 89.90$\pm$2.09 & 87.98$\pm$0.39 & \textbf{95.27$\pm$0.67} \\
		\cdashline{2-12}
		& \textit{avg.} & 89.15$\pm$4.82 & 88.77$\pm$5.43 & 89.83$\pm$2.98 & 89.76$\pm$3.65 & 91.36$\pm$2.7 & 78.87$\pm$10.58 & 83.86$\pm$8.69 & 85.86$\pm$0.81 & \textbf{95.26$\pm$3.10} \\
		\cdashline{2-12}
		& \textit{Rank} & 5 
		& 6
		& 3
		& 4
		& 2
		& 9
		& 8
		& 7
		&1 \\
		\hline
		
		& Benign & 96.29$\pm$1.28 & 91.91$\pm$2.43 & 95.13$\pm$1.62 & 94.42$\pm$1.81 & \textbf{98.47$\pm$0.81} & 89.86$\pm$1.62 & 61.67$\pm$8.09 & 63.38$\pm$1.83 & 96.73$\pm$1.76 \\
		BC& Malignant & 87.68$\pm$2.68 & 84.35$\pm$5.09 & 87.39$\pm$4.90 & 86.57$\pm$4.53 & \textbf{98.36$\pm$1.04} & 62.35$\pm$6.21 & 67.34$\pm$10.38 & 69.85$\pm$1.64 & 96.12$\pm$1.68 \\
		\cdashline{2-12}
		& \textit{avg.} & 91.99$\pm$6.09 & 88.13$\pm$5.35 & 91.26$\pm$5.47 & 90.50$\pm$5.55 & \textbf{98.42$\pm$0.08} & 76.11$\pm$19.45 & 64.51$\pm$4.01 & 66.63$\pm$1.73 & 96.43$\pm$0.43 \\
		\cdashline{2-12}
		& \textit{Rank} 
		& 3
		& 6
		& 4
		& 5
		& 1
		& 7
		& 9
		& 8
		& 2 \\
				
		\hline
		\multicolumn{2}{c} {\textit{Avg. Rank} }
		& 4.3
		&5.0
		& 3.7
		&5.7
		& 1.7
		& 7.7
		& 8.7
		& 7.0
		&1.3 \\
		\hline
	\end{tabular}
	\label{tab4}
\end{table*}

\begin{table}[!h]
	\setlength{\tabcolsep}{3pt} % 默认是6pt，减小这个值
	\centering
	\caption{$p$-values of the Friedman test with post-hoc Wilcoxon signed-rank test for comparisons versus IMD-AD across different dataset types.}
	\scalebox{0.8}{ % 缩放因子
		\begin{tabular}{lcccccccc}
			\hline
			Dataset Type & IF & KDE & OCSVM & SVDD & SSLM & AE & Deep SVDD & DROCC\\
			\hline
			Image Datasets & 0.0000 & 0.0000 & 0.0000 & 0.0000 & 0.0000 & 0.0000 & 0.0000 & 0.0010\\
			Tabular Datasets & 0.0001 & 0.0001 & 0.0001 & 0.0001 & 0.0237 & 0.0001 & 0.0001 & 0.0003\\
			All Datasets & 0.0000 & 0.0000 & 0.0000 & 0.0000 & 0.0000 & 0.0000 & 0.0000 & 0.0000\\
			\hline
			
		\end{tabular}
	}
	\label{tab:friedman_p_values}
\end{table}

To statistically validate the superiority of IMD-AD, we conducted comprehensive significance tests across six benchmark datasets. As summarized in Table~\ref{tab:friedman_p_values}, the Friedman test with post-hoc Wilcoxon signed-rank test demonstrates that IMD-AD significantly outperforms all competing methods ($p < 0.05$), with overwhelming statistical evidence ($p < 0.001$ for most comparisons).

\begin{figure*}[!t]
\centering
\subfloat[Image Datasets]{\includegraphics[width=0.45\textwidth]{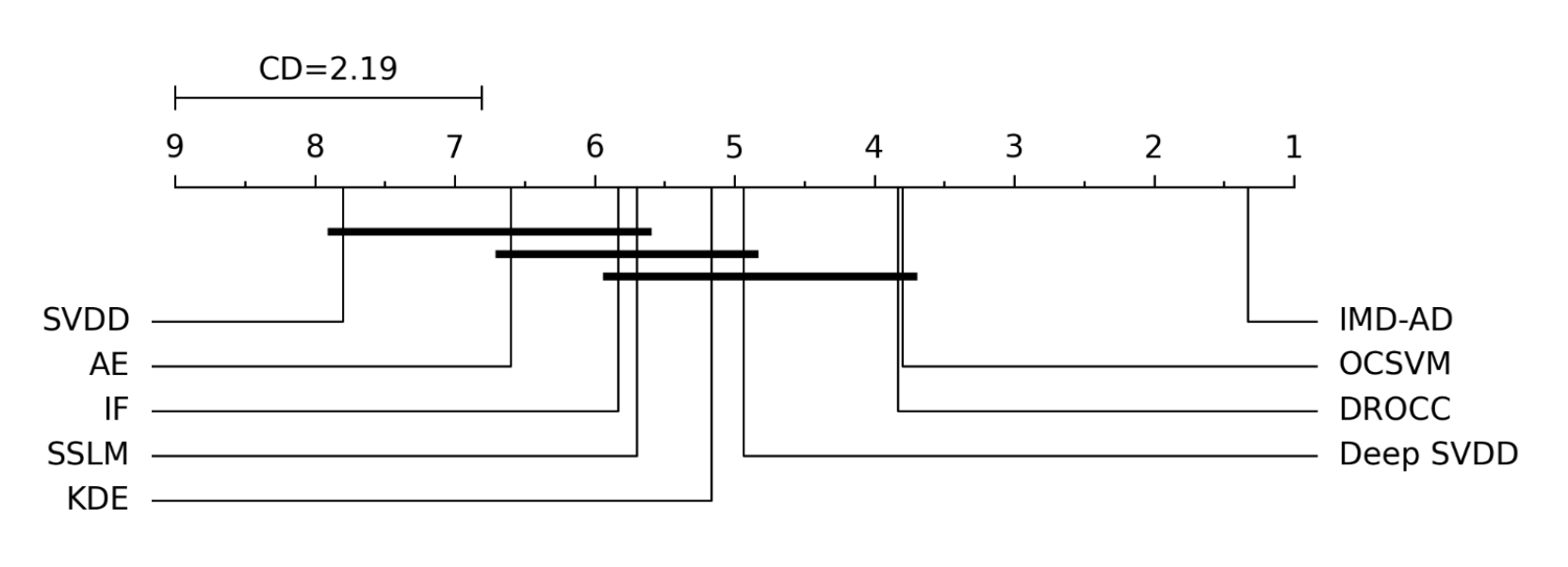}%
\label{fig:cd_image}}
\hfil
\subfloat[Tabular Datasets]{\includegraphics[width=0.45\textwidth]{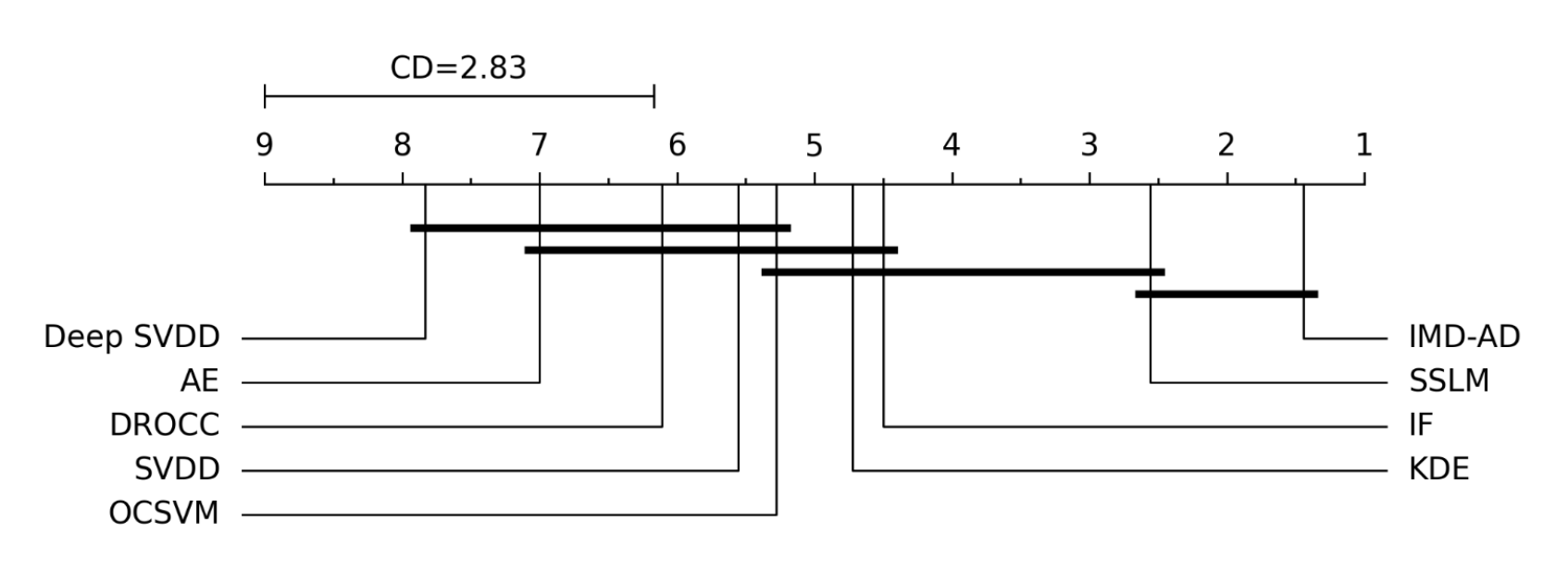}%
\label{fig:cd_tabular}}
\caption{Critical Difference diagrams from Friedman test with Nemenyi posthoc test ($\alpha = 0.05$).}
\label{fig:cd_diagrams}
\end{figure*}

The Critical Difference diagrams in Fig.~\ref{fig:cd_diagrams} further visualize these findings through the Nemenyi posthoc test. IMD-AD achieves the optimal rank in both data categories, with statistically significant separation from other methods. Notably, on image datasets, IMD-AD demonstrates clear superiority over all competitors, while maintaining strong performance on tabular data. These results collectively confirm the robust and consistent advantage of our proposed approach across diverse data modalities.

\subsection{Interpretability Analysis}\label{InterAnaly}
We further analyze the interpretability of the proposed IMD-AD from the aspects of feature extraction and training transparency.

\subsubsection{Example 1 (two synthetic datasets)}
\begin{figure*}[!t]
  \centering
  \includegraphics[width=0.9\textwidth]{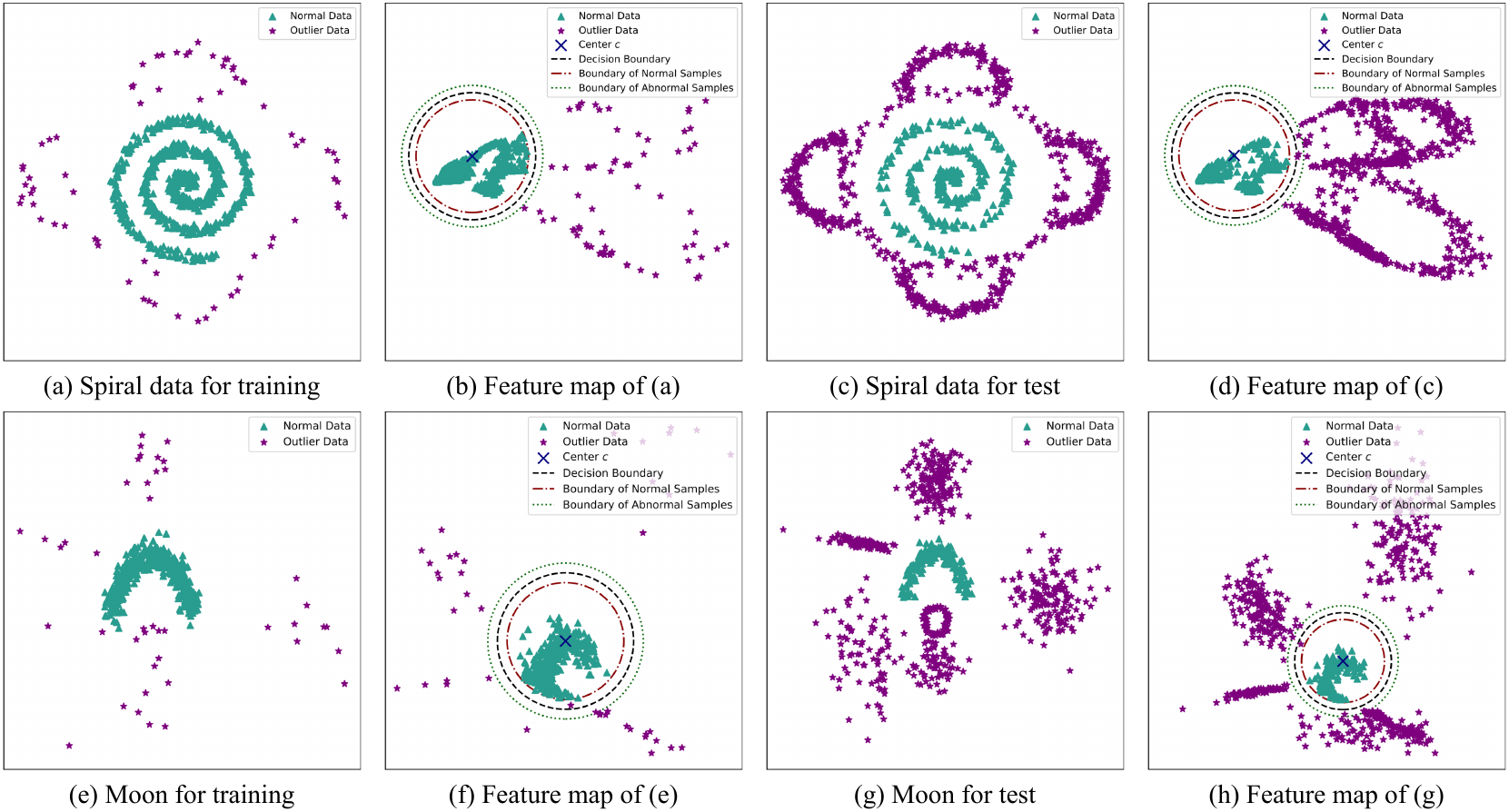}
  \caption{The visualization of IMD-AD on Spiral and Moon Datasets. (a) (c) (e) (g) display the scatter plots on original data. (b) (d) (f) (h) show the extracted samples and the corresponding detection boundaries of IMD-AD.}
  \label{fig3}
\end{figure*}

\begin{figure}[h]
  \centering
  \includegraphics[width=0.35\textwidth]{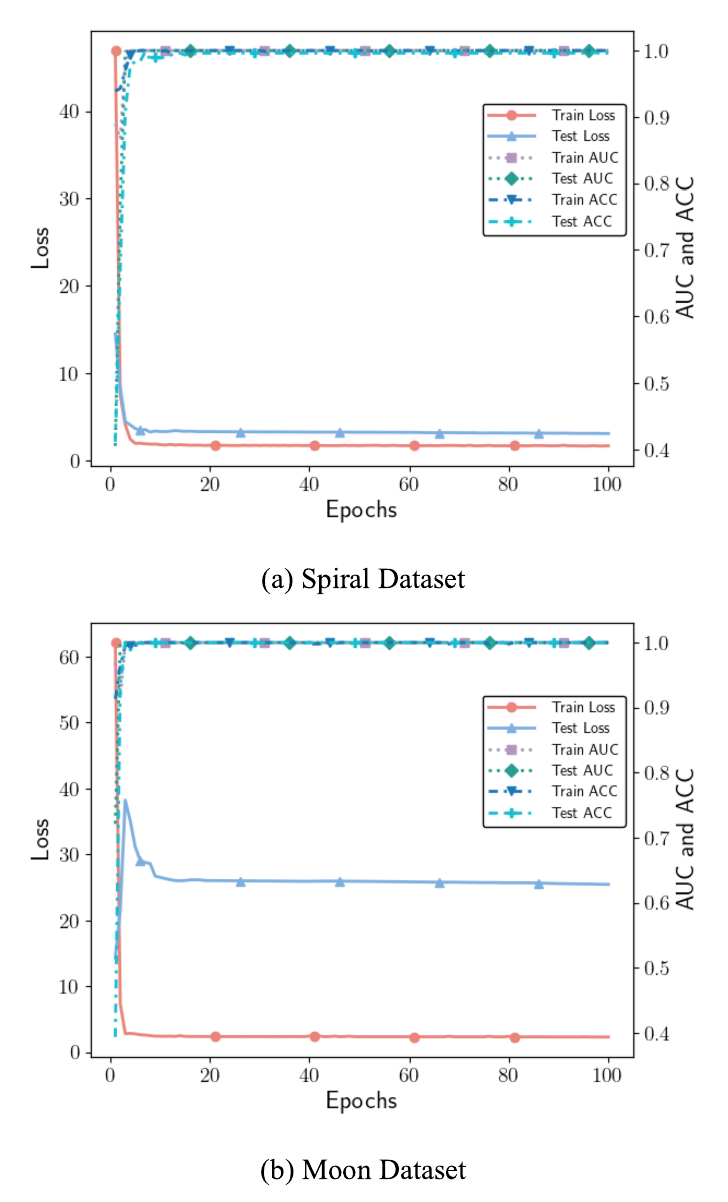}
  \caption{Performance curves of IMD-AD with epochs on Spiral and Moon datasets.}
  \label{fig4}
\end{figure}

We conduct experiments on the Spiral and Moons datasets to illustrate the interpretability of our proposed IMD-AD method.  
Fig.~\ref{fig3} shows the feature extraction performance and classification boundaries for both datasets. Specifically, subfigures (a), (c), (e), and (g) present scatter plots of the original normal and abnormal samples in the input space, while subfigures (b), (d), (f), and (h) depict sample distributions along with the corresponding hypersphere classification boundaries which are obtained when the network neurons $\phi(\mathbf{x}_i; \mathcal{W})$ in Eq. (\ref{eq9}) are fixed to two dimensions.  
As shown in the figures, the inner concentric spheres tightly enclose the normal-class samples, a result of the radius minimization term in our objective function (\ref{eq7}). Meanwhile, the outer concentric boundaries effectively exclude abnormal samples, consistent with our objective of maximizing the margin between normal and abnormal samples in Eq. (\ref{eq7}). This stands in contrast to traditional deep learning approaches, which often lack the ability to produce clearly visualized classification boundaries. Moreover, in our method, both the center and radius of the hyperspheres are learned end-to-end through the neural network.  
Fig.~\ref{fig4} presents the training and testing curves of IMD-AD under different performance metrics, including Accuracy (ACC), AUC, and loss. It can be observed that the model stabilizes after around 40 epochs, accompanied by a significant drop in loss, which confirms the stability and transparency of the training process.

\subsubsection{Example 2 (MNIST dataset)}

\begin{figure*}[!h]
  \centering
  \includegraphics[width=0.9\textwidth]{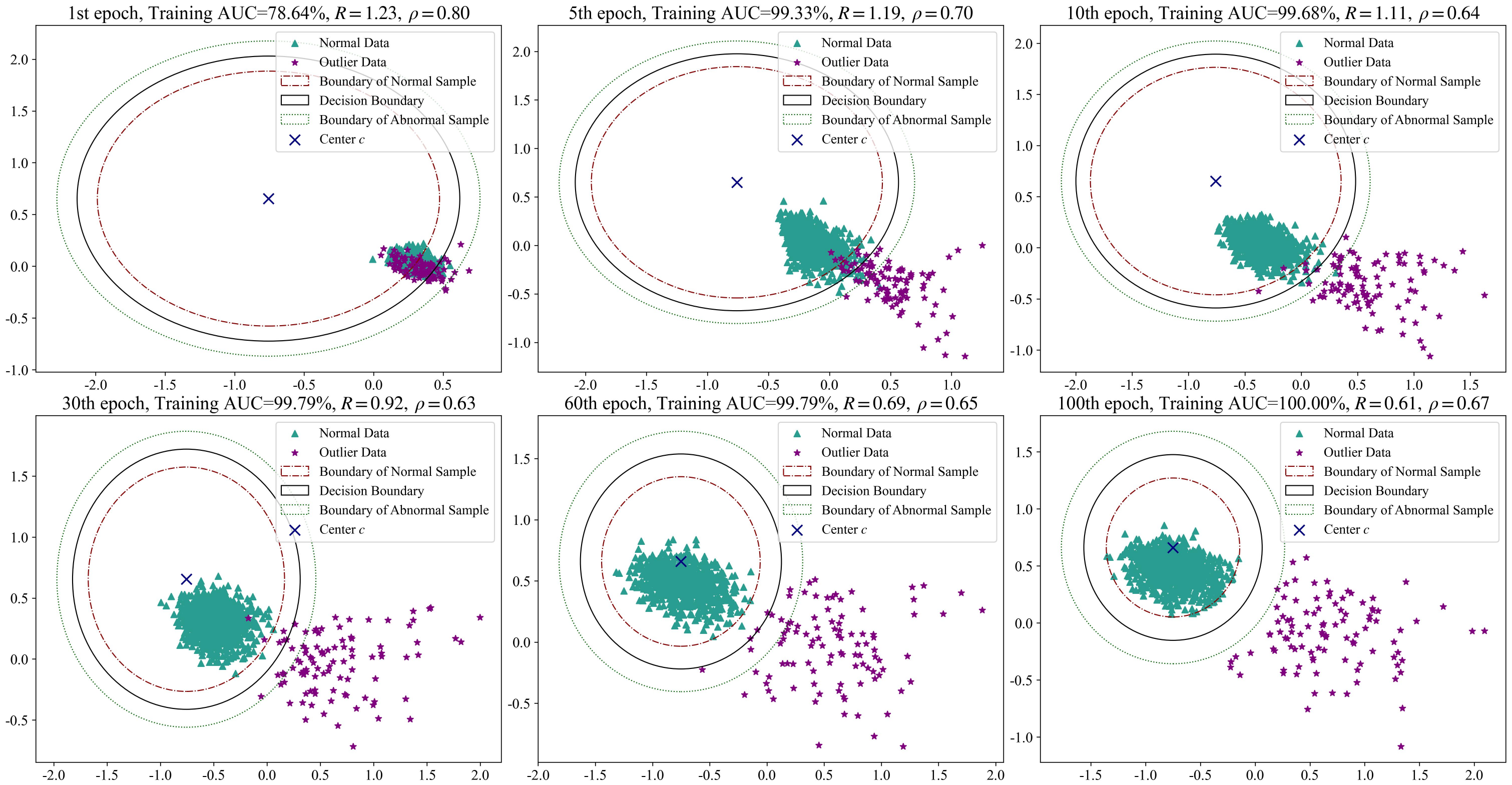}
  \caption{Model classification results on MNIST with normal class 0 over training epochs.}
  \label{fig:mnist_train_epoch}
\end{figure*}

\begin{figure*}[!h]
  \centering
  \includegraphics[width=\textwidth]{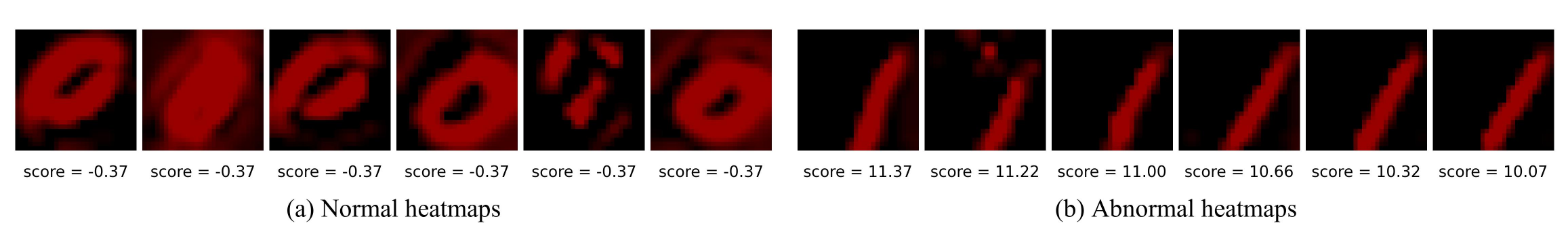}
  \caption{MNIST heatmap visualization comparing normal and abnormal samples for class 0. }
  \label{fig:mnist_heatmap_grid}
\end{figure*}

Fig.~\ref{fig:mnist_train_epoch} presents the classification results on the MNIST dataset across different training epochs, with digit 0 designated as the normal class. As in Example 1, the feature mapping $\phi(\mathbf{x}_i; \mathcal{W})$ is set to be 2-dimensional. At the first epoch, the model has not yet learned effective feature representations, and the distributions of normal and abnormal samples exhibit substantial overlap, leading to a relatively low AUC. From the 5th to the 30th epoch, the model rapidly improves its discriminative ability: normal samples are progressively compressed toward the center of the hypersphere, while abnormal samples are pushed outward. By the 60th epoch, most normal samples are enclosed within the decision boundary and anomalies are clearly separated outside, indicating that the decision function has stabilized. Furthermore, as training progresses, the hypersphere radius $R$ decreases monotonically, reflecting a tighter concentration of normal data around the center, while the margin parameter $\rho$ increases, suggesting enhanced separation between normal and abnormal samples. Although this trend may not be strictly observed in the initial epochs due to random parameter initialization, it becomes consistent as training converges. These results also validate the strong feature extraction capability and training transparency of the proposed IMD-AD, underscoring its intrinsic interpretability.

Fig.~\ref{fig:mnist_heatmap_grid} presents heatmaps for class 0 (digit “0”), which is designated as the normal class in MNIST. Discriminative regions are localized using Grad-CAM (gradient-weighted class activation mapping followed by a ReLU) \cite{selvaraju2017gradcam}, adhering to the CAM localization principle \cite{zhou2016cam}. Let \( s \) denote the anomaly score and \( A^{k} \) the feature map of channel \( k \). The channel weight is defined as \( \vartheta_{k} = Z^{-1} \sum_{i,j} \frac{\partial s}{\partial A^{k}_{ij}} \), where \( Z \) is the number of spatial locations, and the resulting heatmap is given by \( L = \operatorname{ReLU}\!\left( \sum_{k} \vartheta_{k} A^{k} \right) \). To generate normal-class heatmaps, the Grad-CAM objective is set to \( -s \), and its gradient is backpropagated to compute the channel weights and heatmap, thereby emphasizing regions that lower the anomaly score. Conversely, abnormal-evidence heatmaps are obtained using \( +s \), highlighting regions that raise the score. The left panel displays low-scoring normal samples, which predominantly exhibit closed and regular “0”-like shapes. The right panel contains high-scoring abnormal samples, which often resemble the digit “1” with elongated structures that deviate from the typical morphology of “0.” This suggests that the model distinguishes normal from abnormal samples based on structural differences and effectively identifies inputs that diverge from the learned class pattern. In summary, Fig.~\ref{fig:mnist_heatmap_grid} illustrates that, even from a post-hoc explainability standpoint, our proposed method achieves strong performance.

\subsection{Verification of $\nu$-Property}

\begin{figure*}[!h]
  \centering
  \includegraphics[width=\textwidth]{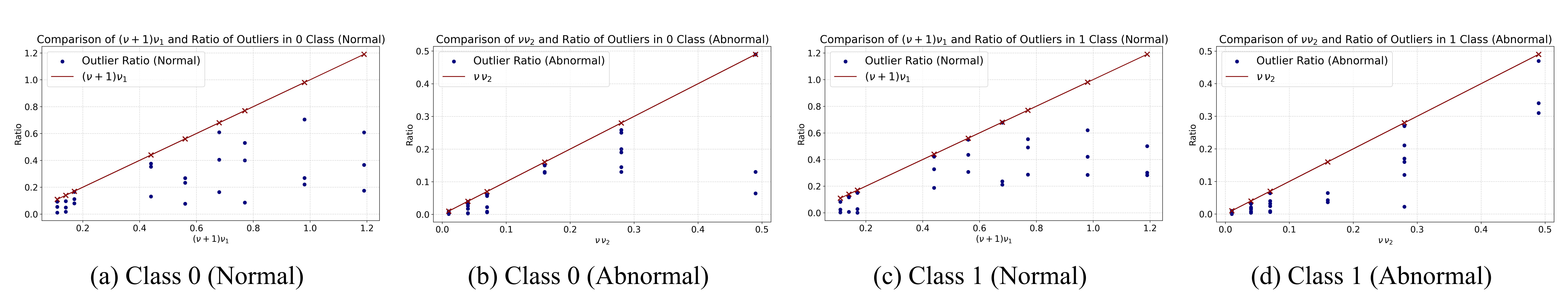}
  \caption{Ratio of outliers with the change of the hyperparameters on the MNIST dataset. The horizontal axis represents the theoretical upper bound determined by the model hyperparameters—$(\nu+1)\nu_1$ for normal data and $\nu\nu_2$ for abnormal data, while the vertical axis reports the empirical outlier ratio under the same settings. In each subfigure, the solid red line depicts the continuous theoretical limit. Red ``$\times$'' markers denote discrete values of that limit. Blue ``$\cdot$'' represent experimental observations. }
  \label{fig:mnist_results}
\end{figure*}

Fig.~\ref{fig:mnist_results} illustrates how the ratio of outliers beyond the classification boundary changes with the hyperparameter on the MNIST dataset. Subfigures (a) and (c) show the results for normal samples from Classes 0 and 1, while the corresponding Subfigures (b) and (d) present the results for abnormal samples of the same classes. Across all subfigures, every blue point lies strictly below its corresponding red ``$\times$'', maintaining a clear margin from the theoretical limit. This pattern holds consistently for all digit classes in both normal and abnormal cases, which directly demonstrate our $\nu$-Property in Section~\ref{vSection}. As the hyperparameters increase from left to right, the theoretical bounds and empirical values rise together, indicating that $(\nu+1)\nu_1$ and $\nu\nu_2$ provide a stable, linear characterization of the variation in outlier ratio. A comparison between normal and abnormal subfigures further shows that abnormal samples exhibit lower overall outlier ratios than normal samples, yet both remain tightly bounded by their respective theoretical limits, underscoring the robustness of these bounds. In summary, Fig.~\ref{fig:mnist_results} provides comprehensive validation of the $\nu$-property for IMD-AD. $(\nu+1)\nu_1$ serves as a reliable upper bound on the outlier rate for normal samples, while $\nu\nu_2$ effectively bounds the outlier rate for abnormal samples.

% \begin{figure*}[!h]
%   \centering
%   \includegraphics[width=\textwidth]{Distance_KDE_merged_four_plots.png}
%   \caption{Empirical density distributions of the squared distances from training samples to the hypersphere center.
% The horizontal axis denotes the squared distance $\|\phi(x)-\mathbf{c}\|^2$ in the feature space, and the vertical axis represents the corresponding empirical density estimated by kernel density estimation.
% The blue dash-dotted curve corresponds to normal training samples (MNIST class~0), while the orange dotted curve corresponds to anomalous training samples (all remaining classes).
% The red dash-dotted vertical line indicates the boundary of normal samples ($\overline{R}$), the black dashed vertical line denotes the decision boundary($T^2$), and the green dotted vertical line represents the boundary of abnormal samples ($\overline{R}+\overline{\rho}$).
% The values shown above the figure report the theoretical upper bounds on the violation ratios, whereas the values listed in the legend correspond to the empirical violation ratios computed on the training set.}
%   \label{fig:mnist_results_KDE}
% \end{figure*}

\begin{figure}[!t]
  \centering
  \includegraphics[width=0.45\textwidth]{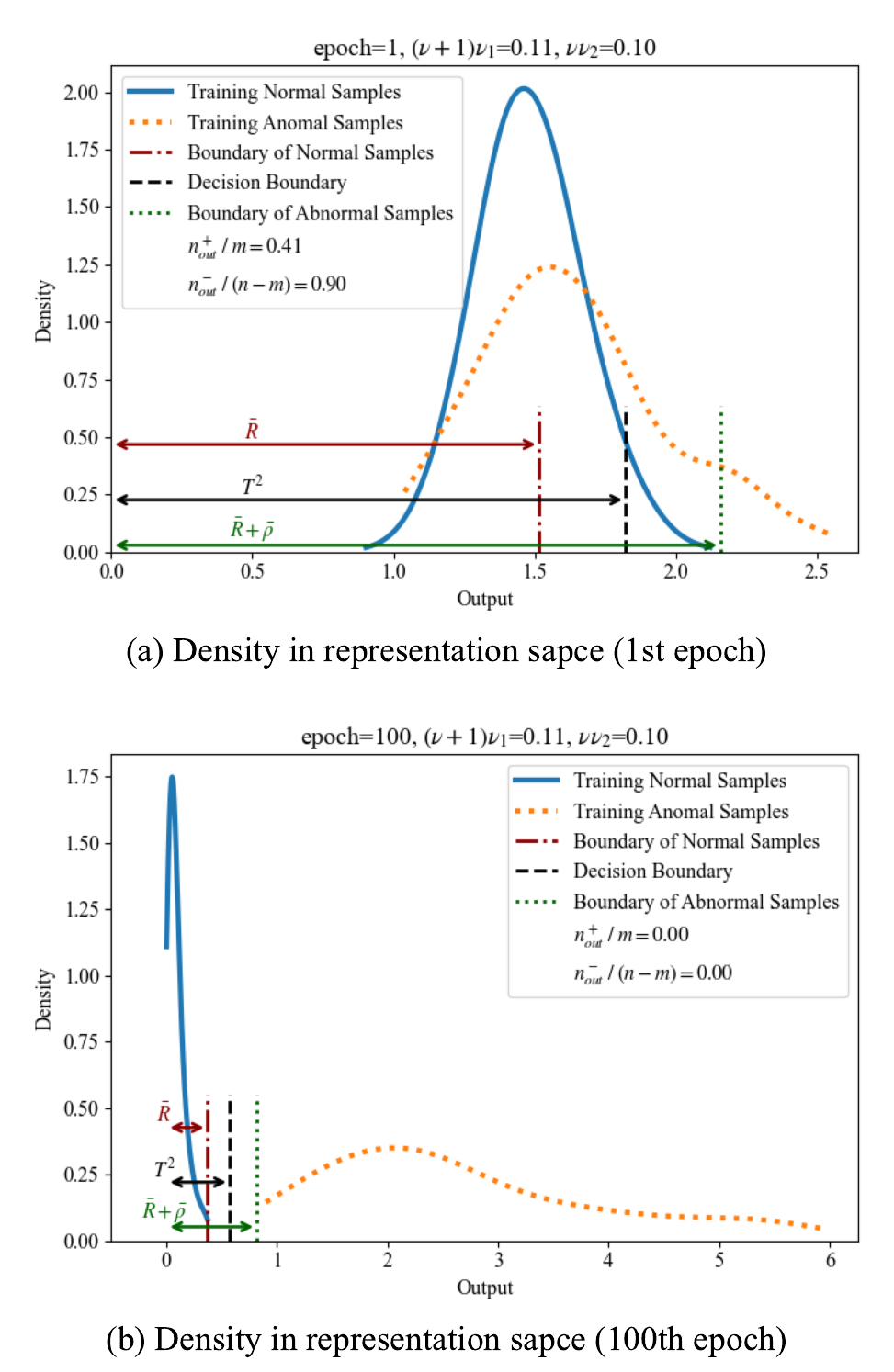}
  \caption{The density of squared distance between training samples and the hypersphere center $\mathbf{c}$ in representation space of MNINST dataset.
Subfigure~(a) shows the distributions at 1st epoch, while subfigure~(b) corresponds to the converged distributions at 100th epoch.}
  \label{fig:mnist_results_KDE}
\end{figure}

Fig.~\ref{fig:mnist_results_KDE} illustrates the density of squared distance between training samples and the hypersphere center $\mathbf{c}$ during training in representation space of MNINST dataset. At the early training stage (1st epoch), the feature representations are not yet discriminative, causing both normal and abnormal training samples to be mapped into a relatively narrow region inside the hypersphere and resulting in highly overlapped distance distributions.
After training converges (100th epoch), the hypersphere parameters and the feature mapping are jointly adapted, leading to a more compact distribution of normal samples with respect to the inner boundary and a relative shift of abnormal samples toward larger distances.
Under these converged conditions, the empirical violation ratios of both normal and abnormal samples are observed to remain below their corresponding theoretical upper bounds.
This observation provides empirical support that the $\nu$-property becomes effective after convergence, with $(\nu+1)\nu_1$ and $\nu\nu_2$ serving as valid upper bounds for the violation ratios in the trained model.

\subsection{Ablation Experiments}
This paper introduces three key improvements over Deep SVDD: incorporating a small number of anomaly samples, introducing the margin parameter $\rho$, and designing a novel interpretable optimization algorithm. To assess the individual contribution of each component to overall model performance, we conduct ablation experiments on the CIFAR-10 dataset. 

The ablation experiments include four methods: 
\begin{itemize}
\item[A.] The vanilla Deep SVDD.
\item[B.] D-AD (Deep Anomaly Detection): only adding a small number of training anomaly samples in the vanilla Deep SVDD.
\item[C.] MD-AD (Maximum Margin Deep Anomaly Detection): adding a small number of anomaly samples and introducing parameter $\rho$ in Deep SVDD, without interpretable neural network parameterization of hypersphere.
\item[D.] IMD-AD: our Interpretable Maximum Margin Deep Anomaly Detection.
\end{itemize}

\begin{figure}[htbp]
	\centering
	% 图片部分（占 50% 行宽）
	\begin{minipage}[t]{0.4\textwidth}
		\raggedright
		\resizebox{1.0\textwidth}{!}{%      
			\begin{tabular}{l c c c | c}
				\hline
				& Abnormal Samples & $\rho$ & Interpretable & Rank \\
				\hline
				Our IMD-AD    &    \checkmark   &     \checkmark  &  \checkmark     & 1 \\				
				MD-AD     &   \checkmark     & \checkmark       &  \ding{55}       &  2\\				
				D-AD   &   \checkmark     &   \ding{55}      &  \ding{55}       &  3\\				
				Deep SVDD  & \ding{55}  & \ding{55} &  \ding{55}  &  4\\
				\hline
			\end{tabular}%       
		}%
	\end{minipage}
    
	\hfill % 填充水平间距
	% 表格部分（占 50% 行宽）
	\begin{minipage}{0.4\textwidth}
		\centering
		\includegraphics[width=1\textwidth]{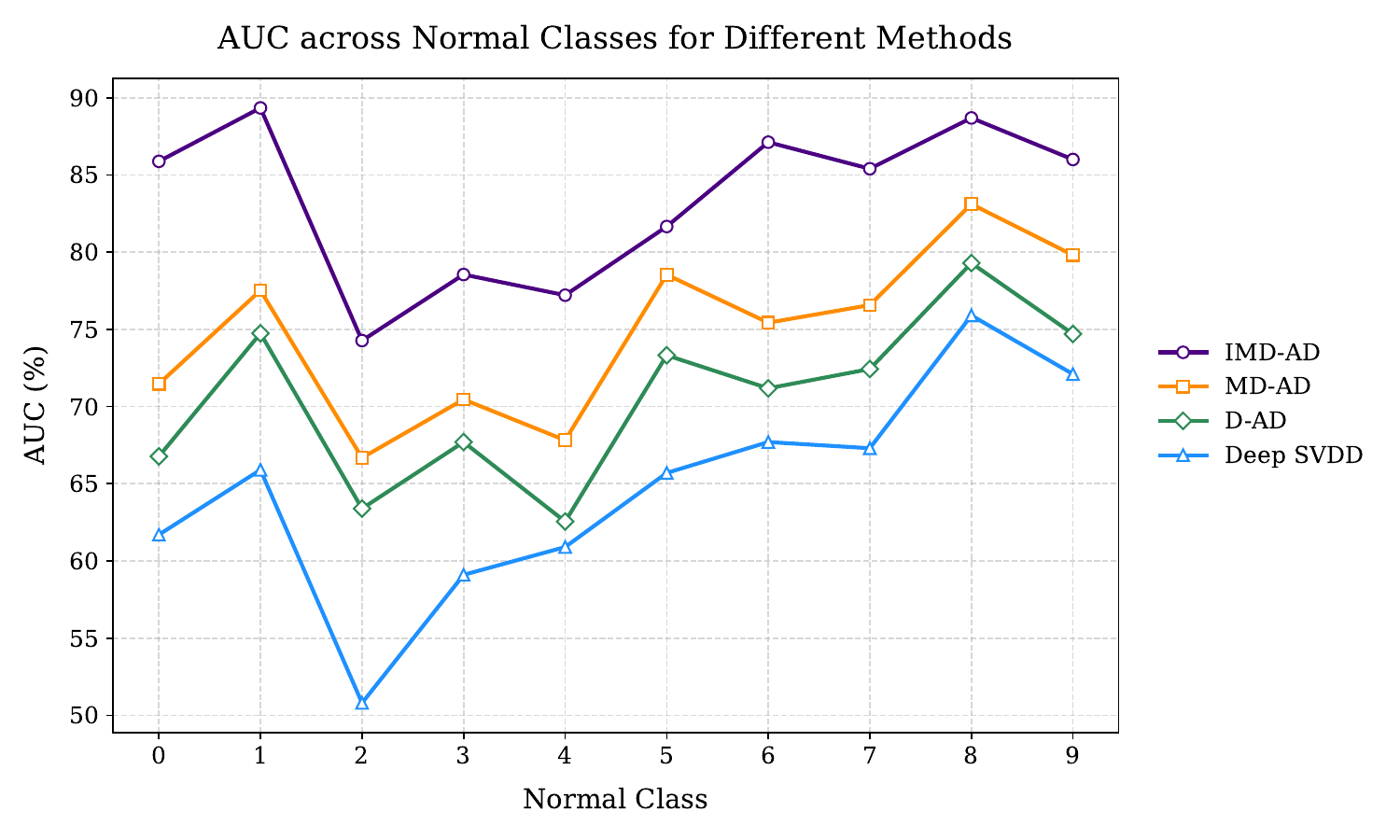}
	\end{minipage}\\		
	\caption{AUC of ablation experiments on the CIFAR-10 dataset.}  
\end{figure}

The main experimental results are presented in Fig.~8. The comparison between Deep SVDD and D-AD indicates that incorporating a small number of anomaly samples during training yields a clear performance improvement. The subsequent transition from D-AD to MD-AD reveals that introducing a margin parameter to refine the decision boundary leads to further performance improvement. Finally, the additional improvement of IMD-AD over MD-AD highlights the effectiveness of our interpretable optimization algorithm. Collectively, these ablation results validate that each proposed component contributes positively to the model performance, while also enhancing interpretability and parameter estimation accuracy.

\section{Conclusion}
\label{ref6}
In this work, we propose Interpretable Maximum Margin Deep Anomaly Detection (IMD-AD). Our method incorporates a small set of abnormal examples into a maximum-margin objective, which both prevents hypersphere collapse and enforces a robust separation between normal and anomalous instances. A key theoretical contribution is the demonstrated equivalence between the hypersphere parameters and the network's final-layer weights. This insight enables joint end-to-end optimization of all model parameters-eliminating reliance on heuristic tuning and improving solution optimality-while naturally providing explicit interpretability through direct visualization and attribution of anomaly scores. Comprehensive experiments on benchmark datasets show that IMD-AD achieves strong performance compared with state-of-the-art methods. By unifying theoretical robustness, end-to-end optimization, and intrinsic interpretability in a single framework, IMD-AD offers a principled and effective advance for deep anomaly detection. Future work will explore extensions to additional anomaly-detection scenarios and further refinement of the interpretability mechanisms.

%\IEEEpubidadjcol

\bibliographystyle{IEEEtran}
% \bibliography{mybib}


\end{document}